%% file: paper.tex
\documentclass[]{fairmeta}

\usepackage{wrapfig}
\usepackage{tabularx}
\usepackage{textcomp}
\usepackage{stfloats}
\usepackage{url}
\usepackage{verbatim}
\usepackage{titlesec}
\usepackage{tocloft}
\usepackage{adjustbox}
\usepackage{multirow}
\usepackage{pifont}
\usepackage{tikz}
\usepackage{comment}
\usepackage{amsmath,amssymb} %
\usepackage{colortbl}  %
\usepackage{color}
\usepackage{booktabs} 
\usepackage{hyperref}
\usepackage{graphicx}    %
\usepackage{subcaption} 
\usepackage{multirow} %
\usepackage{booktabs} %
\usepackage{subcaption} %
\definecolor{headerpurple}{HTML}{d8d2fc}
\RequirePackage{xspace}
\makeatletter
\DeclareRobustCommand\onedot{\futurelet\@let@token\@onedot}
\def\@onedot{\ifx\@let@token.\else.\null\fi\xspace}
\usepackage[most]{tcolorbox}
\usepackage{xcolor}
\usepackage{enumitem}

\newcommand{\huggingface}{\raisebox{-1.5pt}{\includegraphics[height=1.05em]{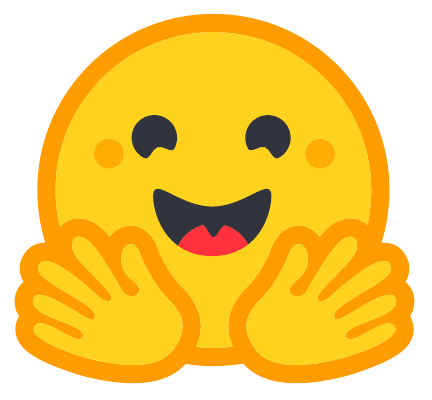}}\xspace}
\newcommand{\github}{\raisebox{-1.5pt}{\includegraphics[height=1.05em]{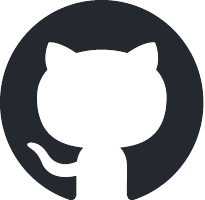}}\xspace}
\newcommand{\homepage}{\raisebox{-1.5pt}{\includegraphics[height=1.05em]{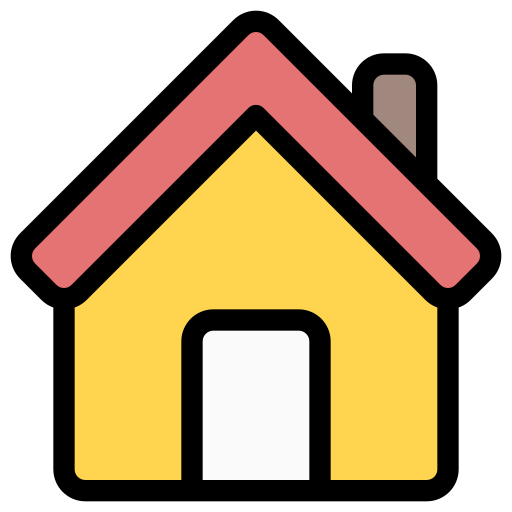}}\xspace}
\newcommand{\modelscope}{\raisebox{-1pt}{\includegraphics[height=0.95em]{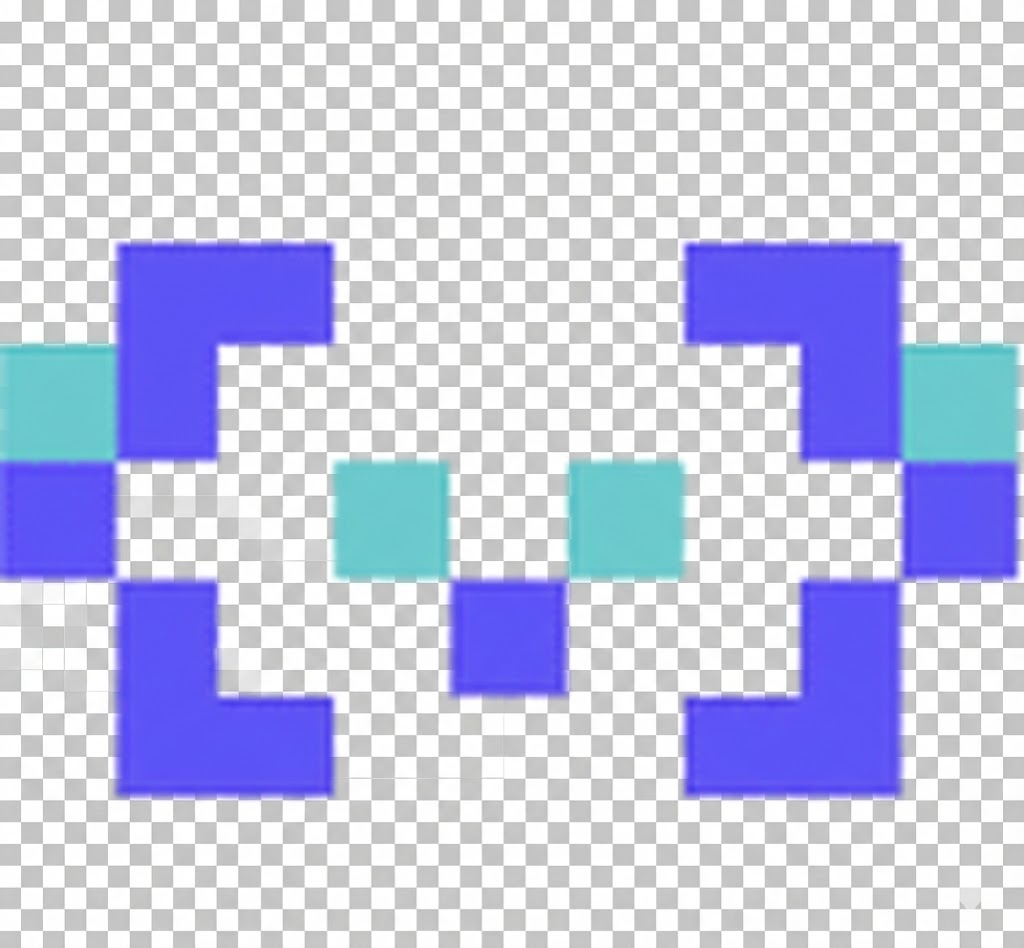}}\xspace}

\makeatother

\definecolor{adptorange}{RGB}{248, 205, 172}
\definecolor{cmpblue}{RGB}{189, 215, 238}

\definecolor{our_red}{RGB}{232,157,160}
\definecolor{our_blue}{RGB}{136,206,230}
\definecolor{our_orange}{RGB}{246,200,168}
\definecolor{our_green}{RGB}{178,211,164}

\definecolor{attn_code0}{RGB}{247,215,200}
\definecolor{attn_code1}{RGB}{238,169,139}
\definecolor{mlp_code0}{RGB}{204,201,221}
\definecolor{mlp_code1}{RGB}{102,95,153}

\definecolor{token_blue}{RGB}{84, 120, 140}

\usepackage{pifont}       %
\usepackage{bbding}       %
\usepackage{fontawesome}

\usepackage{float}
\newcommand{\our}{{RynnWorld-4D}\xspace}
\newcommand{\eg}{\textit{e.g.},\xspace}

\newlength\savewidth

\newcolumntype{x}[1]{>{\centering\arraybackslash}p{#1pt}}
\newcolumntype{y}[1]{>{\raggedright\arraybackslash}p{#1pt}}
\newcolumntype{z}[1]{>{\raggedleft\arraybackslash}p{#1pt}}

\usepackage{colortbl}

\renewcommand{\paragraph}[1]{\vspace{1.25mm}\noindent\textbf{#1}}

\usepackage{algorithm}
\usepackage{listings}

\definecolor{codeblue}{rgb}{0.25, 0.5, 0.5}
\definecolor{codekw}{rgb}{0.35, 0.35, 0.75}
\lstdefinestyle{Pytorch}{
    language = Python,
    backgroundcolor = \color{white},
    basicstyle = \fontsize{9pt}{8pt}\selectfont\ttfamily\bfseries,
    columns = fullflexible,
    aboveskip=1pt,
    belowskip=1pt,
    breaklines = true,
    captionpos = b,
    commentstyle = \color{codeblue},
    keywordstyle = \color{codekw},
}

\definecolor{green}{HTML}{009000}
\definecolor{red}{HTML}{ea4335}

\title{\our: 4D Embodied World Models \\ for Robotic Manipulation}

\author[* 1, 2, 3]{Haoyu Zhao}
\author[* 1]{Xingyue Zhao}
\author[\dagger 1, 4]{Siteng Huang}
\author[1, 4]{Xin Li}
\author[\dagger 1]{Deli Zhao}
\author[\dagger 2, 3]{Zhongyu Li}

\affiliation[1]{DAMO Academy, Alibaba Group,}
\affiliation[2]{Hong Kong Embodied AI Lab,}
\affiliation[3]{CUHK,}
\affiliation[4]{Hupan Lab}

\contribution[*]{Equal contribution}
\contribution[\dagger]{Corresponding author}

\abstract{
Robotic manipulation in the open world requires not only recognizing what a scene looks like, but also anticipating how its 3D structure moves under interaction. We argue that synchronized RGB, depth, and optical flow (RGB-DF) provide a physically grounded representation that captures the underlying 4D dynamics of a scene.
Compared to 2D pixel videos, this multi-modal synergy aligns visual appearance with geometric structure and temporal motion, creating a representation space significantly closer to low-level end-effector actions demanded by robotic systems, narrowing the gap between world prediction and policy learning.
Building on this insight, we introduce \textbf{\our}, a generative model that co-produces future RGB frames, depth maps, and optical flow from a single RGB-D image and a language instruction within one unified diffusion process.
This 4D world model features a tri-branch architecture that integrates cross-modal attention with frame-wise 3D RoPE, ensuring that appearance, geometry, and motion evolve consistently.
To supply training data at scale, we curate \textbf{Rynn4DDataset 1.0}, a massive dataset of over 254.4 million frames across egocentric human and robotic manipulation videos with high-quality pseudo-labels for depth and optical flow.
We further propose \textbf{\our-Policy}, an inverse dynamics head that consumes the internal 4D representations of \our in a single forward pass, bypassing expensive multi-step denoising, to output robot actions in a closed-loop manner. Experiments show that \our produces temporally and spatially coherent 4D predictions, and that \our-Policy achieves state-of-the-art performance on real-world dexterous bimanual manipulation tasks, particularly excelling in tasks demanding spatial precision and temporal coordination.

\begin{center}
    
    \begin{tabular}{ll}
        \homepage  & \url{https://alibaba-damo-academy.github.io/RynnWorld-4D.github.io} \\
        \github  & \url{https://github.com/alibaba-damo-academy/RynnWorld-4D}\\
        \huggingface & \url{https://huggingface.co/Alibaba-DAMO-Academy/RynnWorld-4D} \\
        \modelscope  & \url{https://www.modelscope.cn/models/DAMO_Academy/RynnWorld-4D}\\
    \end{tabular}
\end{center}

}

\date{\today}

\begin{document}
\thispagestyle{firstheader}
\maketitle
\pagestyle{empty}

\input{Sec/1-intro}
\input{Sec/2-related}
\input{Sec/3-method}

\input{Sec/4-exp}
\input{Sec/5-conclusion}

\newpage
\bibliographystyle{assets/plainnat}
\bibliography{paper}

\newpage
\beginappendix
\input{Sec/Appendix}

\end{document}

%% file: Sec/1-intro.tex
\section{Introduction}
Robotic manipulation in the open world could greatly benefit from visual world models that predict how the environment would evolve given an agent’s interactions~\citep{zhao2026towards,zhao2025smap,li2026causal,agarwal2025cosmos,ali2025world}.
While recent generative video models~\citep{ha2018recurrent,xiang2024pandora,zheng2024open,wang2025wan} have shown encouraging progress in policy synthesis~\citep{du2023learning,liang2024dreamitate,zhen2025learning,zhao2026rynnworld}, data simulation and generation~\citep{zhu2024irasim}, and long-horizon planning~\citep{du2023video,li2025novaflow}, they remain limited by the 2D projective nature of pixels. This inherent limitation leads to a loss of critical spatial relationships, preventing precise 6-DoF pose estimation and depth-aware interaction~\citep{hu2024video,agarwal2025cosmos,li2026causal}. Furthermore, 2D models often lack geometric grounding,  leading to temporal inconsistencies such as fluctuating object scales and unphysical shape morphing, which hinders their utility in robust policy learning. Consequently, transitioning generative world modeling from 2D videos to geometry-integrated 4D scene evolution is an essential step toward a solid foundation for embodied intelligence.

Existing 4D scene-modeling approaches fall into two categories. The first builds on Neural Radiance Fields (NeRF)~\citep{mildenhall2021nerf} or 3D Gaussian Splatting (3DGS)~\citep{kerbl20233d}, which can be further divided into optimization-based methods~\citep{zhao2024sg,zhao2024hfgs,yu20244real,bahmani20244d} that are computationally intensive and scene-specific, and feed-forward models~\citep{ren2024l4gm,wu2025cat4d} that prioritize speed but typically focus on object-centric generation. These approaches often require multi-view inputs or struggle to scale to complex scene-level environments. The second category comprises dynamic Structure-from-Motion (SfM) approaches~\citep{wang2025continuous,li2025megasam}, which reconstruct time-varying point clouds but lack the generative capability to predict future states from a single image. Neither category readily provides a compact, scalable representation that integrates with the strong generative priors of pretrained video diffusion models.

\begin{figure}[t]
  \centering
    \includegraphics[width=\linewidth]{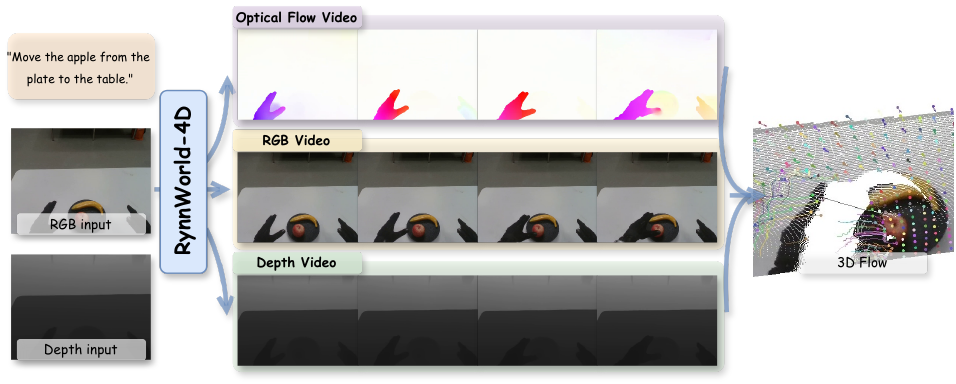}\\
  \caption{Given an input RGB-D image and description, \our generates RGB, depth, and optical flow videos synchronously, which can be further lifted into 3D scene flow (right).}
  \label{fig:teasor}
\end{figure}

To bridge this gap, we propose a lightweight projective 4D representation by predicting synchronized sequences of RGB, depth, and optical flow (RGB-DF): depth lifts each pixel to a 3D location, and depth together with optical flow can be back-projected into 3D scene flow under standard pinhole-camera assumptions, providing a per-point 3D motion cue (illustrated as the ``3D Flow'' in Fig.~\ref{fig:teasor}).
Compared to RGB-only sequences, this representation makes geometry and motion explicit; compared to explicit 3D volumes or 4D Gaussians, it stays in a 2D-aligned format and therefore inherits the scalability and the rich generative priors of large-scale video diffusion models.

Building on this representation, we present \textbf{\our}, a 4D embodied world model that, conditioned on a single RGB-D image and a text instruction, synchronously generates RGB, depth, and optical-flow videos within one shared denoising loop (Fig.~\ref{fig:teasor}).
Specifically, we extend a pretrained video diffusion model~\citep{wang2025wan} into a tri-branch transformer, where each branch handles one modality with independent transformer with shared cross attention keys/values across modalities, while Joint Cross-Modal Attention modules enforce cross-modal consistency.
This design preserves the strong generative priors of the pretrained backbone while allowing each modality to specialize, \textit{i.e.}, textures for RGB, spatial geometry for depth, and motion displacements for optical flow.
A key challenge for training \our is the absence of large-scale datasets with dense 4D annotations. To address this, we curate \textbf{Rynn4DDataset 1.0}, a large-scale hybrid dataset comprising over 254 million video frames drawn from egocentric human activity datasets~\citep{damen2020epic,wang2024egovid} and robotic manipulation datasets~\citep{wu2024robomind,liu2024rdt,jiang2025galaxea,wu2025robocoin,bu2025agibot}, each enriched with high-quality pseudo-annotations for depth and optical flow.

The RGB-DF representation offers a critical advantage: it aligns more closely with a robot’s action space than raw 2D pixel changes. Consequently, a downstream policy trained on \our's internal 4D representations bypasses the heavy structural inference typically required when operating on 2D latents alone. Leveraging this synergy, we introduce \textbf{\our-Policy}, an inverse dynamics head that extracts robot actions directly from \our’s predictive 4D features. By utilizing these internal latents in a single forward pass and bypassing the iterative denoising bottleneck, \our-Policy enables high-frequency, closed-loop control suitable for real-time interaction. In summary, our work makes the following contributions:

\begin{itemize}[labelsep=0.6em, leftmargin=1.2em,itemindent=0em]
\item We introduce a \textbf{projective 4D representation} that co-generates RGB, depth, and optical flow, and we show how it admits a natural 3D-scene-flow reading that makes geometry and motion explicit while staying compatible with large-scale video diffusion priors.
\item We develop \textbf{\our}, a tri-branch 4D world model that co-generates physically coherent RGB-DF sequences through mutual cross-modal interactions.
\item We curate \textbf{Rynn4DDataset 1.0}, a large-scale 4D embodied video dataset with depth and optical flow annotations for training 4D embodied world model.
\item We propose \textbf{\our-Policy}, which leverages the internal 4D representations to enable high-frequency, closed-loop robotic control.
\end{itemize}

%% file: Sec/2-related.tex
\section{Related Work}
\subsection{World Model}
Learning a dynamics model of the world that supports downstream action generation has been a long-standing challenge~\citep{ha2018world,sutton1991dyna}. Early work learns world models in low-dimensional state spaces~\citep{achille2018separation,lesort2018state}, which are efficient to train but difficult to generalize across visually diverse environments. With advances in generative modeling, a growing body of recent work has explored video models as foundation world models~\citep{kong2024hunyuanvideo,wang2025wan,yang2024cogvideox}. However, these models remain in the 2D pixel space, limiting their ability to capture 3D geometric structure and leaving a large representational gap between their predictions and the 3D actions a robot must produce.

3D world models attempt to close this gap by reasoning over meshes or explicit surfaces~\citep{wang2021neus,pfaff2020learning,jiang2025phystwin,zhao2025physsplat,xia2025drawer,xia2024video2game,zhen2025learning,guo2026articulat3d}, radiance fields or Gaussians~\citep{mildenhall2021nerf,kerbl20233d,driess2023learning,xie2024physgaussian}, or particle systems~\citep{sanchez2020learning,abou2024particlenerf,zhang2025particle,chen20254dnex}. Hybrid approaches additionally reason over hierarchical structures~\citep{kaelbling2011hierarchical,wang2025enact,zhao2026high}. While these representations offer richer geometric reasoning, they often require multi-view inputs, are scene-specific, or lack the scalability of pretrained video priors. Most closely related to our work are~\citep{zhen2025learning}, which models the 4D scene from RGB-DN (RGB, Depth, and Normal) videos with language-conditioned control, and~\citep{chen20254dnex}, which produces high-quality dynamic point clouds for novel-view video synthesis.
Our work shares the goal of scalable 4D prediction but introduces a projective 4D representation that co-generates optical flow alongside RGB and depth, making inter-frame 3D motion explicit. Unlike methods relying on static geometry like surface normals~\citep{zhen2025learning}, our inclusion of optical flow allows for back-projection into 3D scene flow, providing explicit dynamic cues essential for learning accurate inverse dynamics. This is particularly critical for dexterous manipulation, where the fine-grained trajectory of objects and end-effectors is what differentiates success from failure.

\subsection{Future Prediction for Embodied Control}
A growing line of work bridges generative modeling and control by using 2D future prediction to guide policy learning~\citep{bharadhwaj2024gen2act,ye2024latent,ye2026world,bi2025motus,hu2024video}. Representative methods include SuSIE~\citep{black2023zero}, which employs a goal-conditioned keyframe generator~\citep{brooks2023instructpix2pix}, and UniPi~\citep{du2023learning}, which learns inverse dynamics over generated sequences. Downstream actions are then derived via online planning~\citep{hu2024video,williams2017model,hafner2019learning,pineau2003point}, offline policy synthesis~\citep{hafner2019dream,hansen2023td,chua2018deep,hafner2025training}, or inverse-dynamics models~\citep{du2023learning,bi2025motus}. However, because these pipelines operate entirely in 2D pixel space and often require repeated denoising for every action step, they face inherent limitations in both geometric accuracy and control reactivity.
In contrast, \textbf{\our} operates on a unified 4D representation that jointly encodes appearance, geometry, and motion. Building upon this, \textbf{\our-Policy} directly consumes the internal predictive features of \our in a single forward pass. By bypassing the need for per-step video decoding and denoising, our approach enables high-frequency, closed-loop robotic control, effectively translating imagined 4D trajectories into precise, real-time robotic actions.

%% file: Sec/3-method.tex
\section{Method}
To address the data scarcity in 4D generative modeling, we first introduce \textbf{Rynn4DDataset 1.0} in Sec.~\ref{sec:data}, a large-scale hybrid dataset specifically curated for training feed-forward 4D generative models. Building upon this, Sec.~\ref{sec:generation} presents \textbf{\our}, a framework capable of co-generating future sequences—including RGB frames, depth maps, and optical flow from a single RGB-D observation and a linguistic task description. Finally, in Sec.~\ref{sec:action}, we introduce \textbf{\our-Policy}, which leverages the predictive 4D representations from \our to derive final robotic actions.

\subsection{Rynn4DDataset 1.0}
\label{sec:data}
\begin{figure}
  \centering
    \includegraphics[width=0.85\linewidth]{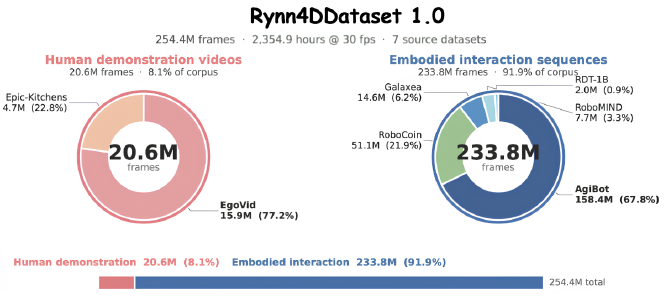}\\
  \caption{\textbf{Composition of the Rynn4DDataset 1.0 dataset.} We provide a large-scale hybrid collection of 254.4M frames, balancing human egocentric videos with diverse robotic manipulation data. This diversity ensures that the world model learns both general object interaction priors and robot-specific execution traces.}
  \label{fig:data-dist}
\end{figure}

\begin{figure}
  \centering
    \includegraphics[width=\linewidth]{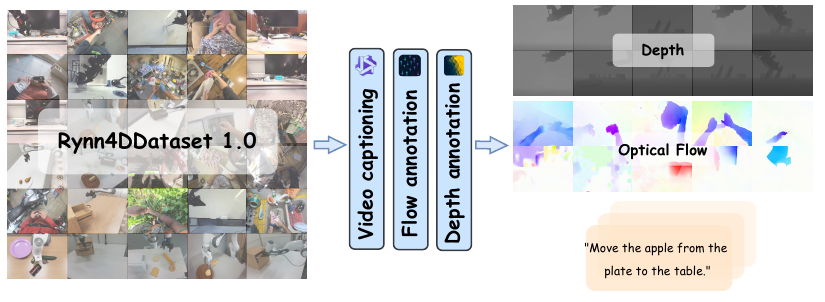}\\
  \caption{\textbf{Data Curation Pipeline.} The video data is collected from diverse sources and partitioned into short clips during data preprocessing. Each clip undergoes a multi-modal annotation process: (1) \textbf{Video Captioning}: Qwen3-VL~\citep{bai2025qwen3} generates
    detailed natural language descriptions of the video content; (2) \textbf{Optical Flow Estimation}: DPFlow~\citep{morimitsu2025dpflow} computes dense per-frame motion fields, which are visualized and saved as flow videos; (3) \textbf{Depth Estimation}: Depth Anything
   3~\citep{lin2025depth} produces monocular depth predictions, which are upsampled to the original resolution and saved as depth videos with a global depth range of $[0.0, 5.0]$ meters.}
  \label{fig:data}
\end{figure}

To bridge the gap in large-scale 4D training data, we introduce Rynn4DDataset 1.0, a hybrid dataset comprising over 254.4 million video frames from human-centric (Epic-Kitchens~\citep{damen2020epic}, EgoVid~\citep{wang2024egovid}) and robotic manipulation datasets (RoboMIND~\citep{wu2024robomind}, RDT-1B~\citep{liu2024rdt}, Galaxea~\citep{jiang2025galaxea}, RoboCoin~\citep{wu2025robocoin}, AgiBot~\citep{bu2025agibot}).
Each frame is enriched with high-quality 4D pseudo-annotations: fine-grained instructions~\citep{bai2025qwen3}, monocular depth~\citep{lin2025depth}, and dense optical flow~\citep{morimitsu2025dpflow}.
The statistics and composition of Rynn4DDataset 1.0 are visualized in Fig.~\ref{fig:data-dist}.
The pipeline for our multimodal annotation is illustrated in Fig.~\ref{fig:data}.

\noindent
\textbf{Video captioning.}
We use Qwen3-VL~\citep{bai2025qwen3} to generate captions for the video data~\citep{damen2020epic,wang2024egovid,wu2024robomind,liu2024rdt,jiang2025galaxea}. Specifically, we leverage the model's strong video-language understanding capabilities to produce detailed, structured descriptions of each video clip. The videos are first sampled at a frame rate of 1 FPS and split into segments of 5 seconds. For each segment, we provide the following prompt to the model:

\begin{quote}                                                                                    \texttt{Please describe this video in detail. Include the following aspects:} \

\texttt{1. The main subject and action in the video.} \

\texttt{2. The environment and background.} \

\texttt{3. Any objects and their interactions.} \

\texttt{4. The overall scene context and atmosphere.} \

\texttt{Provide a concise but comprehensive caption in one paragraph.}                           \end{quote}

The generation is performed with a maximum output length of 512 tokens and a temperature of 0.7 to balance creativity and coherence. The generated captions are then collected and stored in JSON format for downstream tasks.

\noindent
\textbf{Optical flow annotation.}
We employ DPFlow~\citep{morimitsu2025dpflow}, a state-of-the-art optical flow estimation model. For each video, frame pairs are processed sequentially at native resolution, and the estimated flow fields are visualized via color encoding and saved as MP4 videos at 25 FPS.

\noindent
\textbf{Depth annotation.}
We employ Depth Anything 3~\citep{lin2025depth}, specifically the \texttt{DA3NESTED-GIANT-LARGE-1.1} checkpoint, which provides dense per-frame depth predictions along with camera pose estimation. Each video is sampled at 30 FPS and processed at a working resolution of 392 pixels (short side, upper-bound resize).

To convert the estimated depth maps into viewable depth videos, we load the compressed depth arrays and upsample each frame to the original video resolution using bilinear interpolation. Depth values are clipped to a global range of $[0.0, 5.0]$ meters and quantized to 8-bit grayscale via $I = \lfloor d / d_{\max} \times 255 \rfloor$. The resulting frames are saved as RGB videos.

\subsection{3D Scene Reconstruction from Multi-Modal Videos}
\label{sec:3d_reconstruction}

A key advantage of the RGB-DF (\textit{i.e.}, RGB, depth, and optical flow) representation is its inherent geometric interpretability. By combining the co-generated depth and optical flow, we can reconstruct a temporally consistent 3D scene and derive \textit{metric scene flow}.

\noindent
\textbf{Geometric Unprojection.}
Given the generated depth map $D_t$ at frame $t$, each pixel $\mathbf{p}_t = [u, v, 1]^\top$ in homogeneous coordinates is unprojected into the 3D camera space as:
\begin{equation}
    \mathbf{P}_t = D_t(u, v) \cdot \mathbf{K}^{-1} \mathbf{p}_t,
\end{equation}
where $\mathbf{K}$ is the camera intrinsic matrix. This process lifts the 2D projective sequence into a metric 3D point cloud $\mathcal{C}_t = \{\mathbf{P}_t^i\}_{i=1}^{H \times W}$.

\noindent
\textbf{Metric Scene Flow Derivation.}
To capture the underlying 4D dynamics, we leverage the co-generated dense optical flow $\mathbf{f}_{opt} = [\Delta u, \Delta v]^\top$ to establish temporal correspondences. A 3D point $\mathbf{P}_t$ is tracked to its position at $t+1$ by:
\begin{equation}
    \mathbf{P}_{t+1} = D_{t+1}(u+\Delta u, v+\Delta v) \cdot \mathbf{K}^{-1} (\mathbf{p}_t + [\Delta u, \Delta v, 0]^\top).
\end{equation}
The \textit{3D scene flow} is then defined as $\mathbf{f}_{3D} = \mathbf{P}_{t+1} - \mathbf{P}_t$, representing the per-point metric displacement. This explicit 4D mapping ensures that the generated trajectories are not merely visual hallucinations but correspond to physically plausible 3D movements.

\noindent
\textbf{Refinement and Visualization.}
To suppress artifacts at depth discontinuities, we apply a depth-gradient-based edge filter, masking out pixels where $\|\nabla D\| > \tau$. The resulting refined 3D trajectories are projected into a canonical bird's-eye view (BEV). In our qualitative analysis (see the ``3D Flow'' panel in Fig.~\ref{fig:teasor}), these trajectories are rendered as colored trails over a depth-ordered point cloud backdrop, providing an intuitive verification of the model's spatial-temporal coherence.

\begin{figure}[t]
  \centering
    \includegraphics[width=\linewidth]{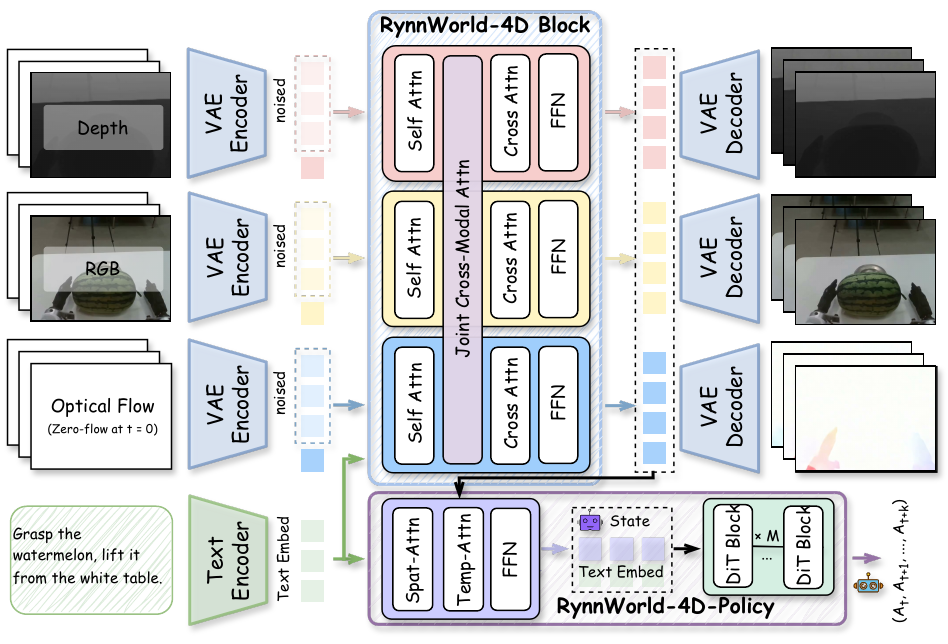}\\
  \caption{\textbf{Overview of \our.} Our pipeline leverages the large-scale Rynn4DDataset 1.0 dataset to train a generative model capable of predicting future 4D sequences. Given a single RGB-D observation and a language instruction, \our co-generates future RGB frames, depth maps, and optical flow. These predictive 4D representations are then aggregated by \our-Policy to derive the final robot actions.}
  \label{fig:pipeline}
\end{figure}

\subsection{\our}
\label{sec:generation}
To achieve synchronized generation of RGB-DF sequences, we extend a pretrained video generative model into a tri-branch architecture (see the overview in Fig.~\ref{fig:pipeline}). This representation is not merely a concatenation of channels; it admits a physically-grounded unprojection into 3D scene flow, as detailed in Sec.~\ref{sec:3d_reconstruction}.
We denote the latents for modality $m \in \{\text{rgb, depth, flow}\}$ as $\bm{z}_t^m$, where $t \in [0, 1]$ represents the flow-matching timestep. Each $\bm{z}_t^m \in \mathbb{R}^{T \times C \times H \times W}$ encapsulates the entire temporal sequence of $T$ frames.

\noindent
\textbf{Tri-branch Architecture.}
To inherit the powerful generative priors of the pretrained model while capturing the distinct characteristics of each modality, we expand the single-branch backbone of Wan~\citep{wang2025wan} into a tri-branch structure.
This decoupled design allows each modality to model its unique feature distributions, such as complex textures for RGB, spatial geometry for depth, and motion displacements for flow to mitigate representation interference among divergent modalities.

\noindent
\textbf{Joint Cross-Modal Attention.}
To enforce cross-modal consistency, we introduce a Joint Cross-Modal Attention (JA) module that is inserted every \emph{three} transformer blocks across all 30 Wan-2.2 layers (at layers $0, 3, 6, \dots, 27$), yielding 10 JA modules in total. Each JA module is appended after the intra-modal self-attention of its host block.

Before cross-modal mixing, each branch $m \in \{\text{rgb,depth,flow}\}$ receives a learnable modality embedding $\bm{e}^m \in \mathbb{R}^{1\times 1\times d}$ (zero-initialized so the module starts as a pure residual) and is normalized by a per-modality LayerNorm $\operatorname{LN}^{m}$ to align numerical scales across branches:
\begin{equation}
\tilde{\bm{z}}_l^m = \operatorname{LN}^{m}\bigl(\bm{z}_l^m + \bm{e}^m\bigr).
\end{equation}

Each branch $m$ produces \emph{one} query and \emph{one} shared key/value pair that is reused by all other branches' queries, reducing the parameter cost from $18d^2$ to $12d^2$ per block:
\begin{equation}
\bm{Q}_l^m = \operatorname{RMSNorm}_q\!\bigl(\operatorname{QProj}_l^m(\tilde{\bm{z}}_l^m)\bigr),\quad
[\bm{K}_l^m, \bm{V}_l^m] = \operatorname{KVProj}_l^m(\tilde{\bm{z}}_l^m),\quad
\bm{K}_l^m \!\leftarrow\! \operatorname{RMSNorm}_k(\bm{K}_l^m).
\end{equation}

Tokens are reshaped from $[B, T{\cdot}S, d]$ to $[B{\cdot}T, S, d]$ so that cross-modal attention is restricted to tokens of the \emph{same temporal frame} across modalities, and 3D Rotary Positional Embeddings are applied to $\bm{Q}_l^m$ and $\bm{K}_l^m$ to inject spatial position information consistently across branches. Each query attends only to the keys/values of the two complementary modalities:
\begin{equation}
\bm{A}_l^m = \operatorname{Attn}\bigl(\operatorname{RoPE}(\bm{Q}_l^m),\;\operatorname{RoPE}(\bm{K}_l^{\text{cross}}),\;\bm{V}_l^{\text{cross}}\bigr),             \end{equation}
with $\bm{K}_l^{\text{cross}} = \operatorname{concat}(\{\bm{K}_l^j\}_{j\neq m})$ and $\bm{V}_l^{\text{cross}} = \operatorname{concat}(\{\bm{V}_l^j\}_{j\neq m})$.

Instead of the double zero-initialization used in ControlNet—which we found to introduce a saddle-point deadlock—we combine a \emph{zero-initialized} output projection $\operatorname{OutProj}_l^m$ with a learnable gate $g_l^m$ initialized to $1$:
\begin{equation}
\hat{\bm{z}}_l^m = \bm{z}_l^m + \tanh(g_l^m)\cdot \operatorname{OutProj}_l^m(\bm{A}_l^m).                                \end{equation}

At initialization $\operatorname{OutProj}_l^m \equiv 0$ guarantees a smooth warm start from the Stage-1 checkpoint, while $\tanh(g_l^m) = \tanh(1) \neq 0$ ensures non-zero gradients flow into the gate so that it can decrease, increase, or change sign as training proceeds, preventing the joint pathway from being trapped at the origin.

\noindent
\textbf{Phased Training Strategy.}
To bridge the significant distribution gaps between modalities, we propose a phased training paradigm:
\textbf{Stage 1: Modality Adaptation.} In this initial stage, we disable the Joint Cross-Modal Attention and train the three branches independently. This allows the depth and flow branches to effectively adapt to their respective geometric and kinetic distributions.
\textbf{Stage 2: Joint Attention Training.} We insert Joint Cross-Modal Attention modules every three layers across all 30 transformer blocks. The entire backbone and per-branch self-attention/FFN are frozen; only the Joint Cross-Modal Attention projections, RMSNorms, per-modality LayerNorms, tanh gates, and the three modality embeddings are trainable. Joint Cross-Modal Attention uses 3D RoPE and a frame-wise mask so cross-modal attention stays within the same temporal frame.
\textbf{Stage 3: Full-Parameter Joint SFT.} With the joint module already aligned, we unfreeze the entire model and continue on the full Rynn4DDataset 1.0.

\noindent
\textbf{Branch Dropout.}
In Stages 2 and 3, with probability $p_{\text{drop}}$ we randomly select one of $\{\text{depth}, \text{flow}\}$ at each training step and replace its noisy latent (frames $[1{:}]$) with pure Gaussian noise, forcing the JA modules to reconstruct it from the visible modalities. The RGB branch is never dropped, since it serves as the appearance anchor: destroying it would leave the joint module with no consistent reference.

\noindent
\textbf{Training Objective.}
All three stages are optimized using the flow matching objective~\citep{lipman2022flow}. For each modality $m \in \mathcal{M} = \{\text{rgb}, \text{depth}, \text{flow}\}$, we learn a velocity field $\bm{v}_\theta^m$ that transports Gaussian noise $\bm{\epsilon}^m$ to data $\bm{z}_0^m$ along the path $\bm{z}_t^m = (1-t)\bm{z}_0^m + t\bm{\epsilon}^m$. The first frame of each modality is the clean image-to-video conditioning latent (a real RGB frame, a real depth frame, and a zero-flow frame, respectively) and is excluded from supervision; we use the slice $[1{:}]$ to denote frames $1,\dots,T-1$. The total loss is:
\begin{equation}
\label{eq:flow_matching_main}
\mathcal{L}_{\text{total}} \;=\; \sum_{m \in \mathcal{M}} \lambda_m \,
        \mathbb{E}_{\bm{z}_0^m, \bm{\epsilon}^m, t, \bm{c}}
        \!\left[\bigl\| \bm{v}_\theta^m\!\bigl(\bm{z}_t^m, t, \bm{c}\bigr)_{[1{:}]}
        - \bigl(\bm{\epsilon}^m - \bm{z}_0^m\bigr)_{[1{:}]} \bigr\|_2^2\right],
\end{equation}
where $\bm{c}$ denotes the text prompt together with the initial RGB-D observation, and $\bm{\epsilon}^{\text{rgb}} = \bm{\epsilon}^{\text{depth}} = \bm{\epsilon}^{\text{flow}}$ is a single Gaussian noise sample shared across the three modalities so
that their denoising trajectories stay temporally aligned. The modality weights are $\lambda_{\text{rgb}} = \lambda_{\text{depth}} = 1$ throughout, while $\lambda_{\text{flow}} = 0.5$ in Stage 1 (the flow first frame carries no informative signal at warm-up) and $\lambda_{\text{flow}} = 1.0$ in Stages 2 and 3.

\subsection{\our-Policy}
\label{sec:action}
We leverage \our as a \textit{predictive 4D vision encoder}.
Given the current RGB-D observation and instruction, we perform a forward pass through the frozen \our, which yields a latent trajectory encoding future dynamics—serving as a powerful representation for robotic manipulation.
We extract intermediate hidden states across all branches and concatenate them along the channel dimension to form $F_p \in \mathbb{R}^{B \times T \times 3C \times H \times W}$, where $C$ is the latent channel dimension per branch. By concatenating features from these decoupled branches, \our-Policy benefits from specialized representations: the RGB branch provides rich visual context, while the independent depth and flow branches offer explicit geometric and kinetic cues, respectively.

\noindent
\textbf{Flow Former.}
To compress 4D features, we use a Flow Former with learnable queries $\bm{Q}$. It processes $F_p$ via frame-wise spatial cross-attention followed by temporal self-attention:
\begin{equation}
 \bm{Q}'_i = \operatorname{Spat-CrossAttn}(\bm{Q}_i, F_p[i]), \quad \bm{Q}'' = \operatorname{FFN}(\operatorname{Temp-SelfAttn}(\bm{Q}')), i \in \{1, \dots, T\}
\end{equation}
where $i$ indexes the frame sequence, and $\bm{Q}''$ encapsulate the predicted spatio-temporal dynamics.

\noindent
\textbf{Action Generation.}
We employ a flow matching~\citep{lipman2022flow} policy to generate actions, following the objective defined in Eq.~\ref{eq:flow_matching_main}. Here, the velocity field $v_\phi$ operates on the action space $\bm{a}$, conditioned on the predictive 4D tokens $\bm{Q}''$, text embedding $l_{\text{emb}}$, and proprioception $p_0$. During inference, the action is derived via an ODE solver in $N=4$ steps, enabling high-frequency closed-loop control through parallel action chunking (see Sec.~\ref{appendix:latency} for details).

\subsection{Inference Latency and Real-time Control}
\label{appendix:latency}
To evaluate the feasibility of \our-Policy in real-world scenarios, we conduct a detailed timing analysis of our inference pipeline. The model is deployed on a workstation equipped with an NVIDIA RTX 5090 GPU, leveraging FP8 quantization and FlashAttention 3 (FA3) kernels to accelerate the transformer-based 4D generation.

\textbf{Note that the 4D visual features are extracted from the frozen \our in a single forward pass ($N=1$). The subsequent 4-step ODE sampling for action generation occurs only within the lightweight \our-policy head.}

\noindent
\textbf{Latency Breakdown.}
The overall control frequency is determined by the total cycle time of the \our\ forward pass and the action generation head. Given a sequence of $K=10$ actions generated per forward pass, a control frequency of approximately 9~Hz is achieved with a cycle time of $\sim$1.1~s. Tab.~\ref{tab:latency_breakdown} provides a granular breakdown of the time spent in each phase.

\begin{table}[t]
\centering
\caption{Inference latency breakdown on NVIDIA RTX 5090 (FP8).}
\label{tab:latency_breakdown}
\begin{tabular}{@{}lccc@{}}
\toprule
\textbf{Phase} & \textbf{Latency (ms)} & \textbf{Percentage}  \\
\midrule
\rowcolor{gray!10}Depth Estimation (DA3 \cite{lin2025depth}) & 85 & 7.7\%  \\
VAE Encoding \& Latent Prep & 18 & 1.6\%  \\
\rowcolor{gray!10}\our & 990 & 89.5\%  \\
Feature Reshape \& Concat & 1 & 0.1\%  \\
\rowcolor{gray!10}Flow Former & 4 & 0.4\%  \\
Action Flow Matching Head & 8 & 0.7\% \\
\rowcolor{headerpurple!60} \textbf{Total Evaluation (Forward)} & \textbf{1,106} & \textbf{100\%}  \\ \bottomrule
\end{tabular}
\end{table}

The primary bottleneck is the tri-branch Transformer, which accounts for 89.5\% of the total latency.

\noindent
\textbf{Control Frequency.}
It is important to distinguish between the \textit{planning frequency} (the rate at which the 4D world model refreshes its mental state) and the \textit{effective control frequency}. Although a single forward pass takes $\sim$1.1~s (yielding a planning frequency of $\approx$0.9~Hz), the \our-Policy employs \textbf{action chunking} by predicting $K=10$ future steps in a single inference. As these 10 actions are executed sequentially while the next planning cycle is computed in parallel, the system achieves an effective control frequency of $\approx$9~Hz.

\noindent
\textbf{Closed-loop Robustness.}
While 9~Hz is lower than traditional low-level PID controllers (typically $>$500~Hz), \our-Policy maintains high robustness through two mechanisms:

Instead of a single action, the policy outputs a sequence of $K=10$ future actions. During the $\sim$1.1s inference cycle, the robot executes the previously planned action chunk at 50~Hz via a cached lookup. The 9~Hz update rate is sufficient to capture most human-scale manipulation dynamics.

Unlike 2D policies that suffer from visual aliasing or depth ambiguity, our policy's internal latents are grounded in 3D scene flow. As shown in our ablation, the inclusion of explicit kinetic cues allows the policy to predict object movements. This anticipation compensates for the slight sensing-to-actuation lag, as the model is not just reacting to the current frame but is conditioned on a predicted 4D trajectory.

In real-world tests, we observe that even when objects are slightly bumped during the 1s execution window, the next 9~Hz update effectively re-plans the trajectory because the RynnWorld-4D latents encompass a spatial volume rather than just a pixel point, providing a wider capture range for recovery.

%% file: Sec/4-exp.tex
\section{Experiments}
\subsection{Implementation Details}
\label{sec:implementation}

\subsubsection{\our}
Our \our\ model is built upon the Wan 2.2-TI2V-5B diffusion transformer~\citep{wang2025wan}, a 30-layer DiT with hidden dimension $d=3072$ and FFN dimension $14{,}336$. We extend its native single-branch RGB backbone into a unified tri-branch architecture for the synchronous synthesis of RGB, depth, and optical flow sequences. The depth and flow branches are initialized by duplicating the pre-trained components—patch embeddings, self-attention, normalization layers, and FFNs—leveraging the robust spatial-temporal priors of the video backbone.

To ensure semantic alignment while minimizing overhead, we share the text cross-attention Key/Value projections across all three branches, as the linguistic task description provides a modality-agnostic semantic signal. We insert Joint Cross-Modal Attention (JA) modules every three transformer blocks across all 30 layers (at layers 0, 3, 6, \dots, 27), yielding 10 modules in total. For each branch $m$, the JA module queries the concatenated K/V pairs from the complementary modalities $j \neq m$:
\begin{equation}
    \hat{\bm{z}}_l^m = \bm{z}_l^m + \tanh(g_l^m) \cdot \operatorname{CrossBranchAttn}(\bm{Q}_l^m, \bm{K}_l^{\text{cross}}, \bm{V}_l^{\text{cross}}),
\end{equation}
where $g_l^m$ is a learnable scalar gate initialized to $1$. To ensure a smooth transition from independent branch training, we initialize the output projection to zero while keeping $g_l^m=1$. JA employs 3D RoPE and a frame-wise mask to restrict attention to tokens within the same temporal frame.

\paragraph{Phased Training Strategy.}
We adopt a three-stage curriculum to bridge modality distribution gaps. Tab.~\ref{tab:training_hparams} summarizes the stage-wise configuration. Each stage is initialized from the \emph{model-only} checkpoint of the previous stage (optimizer/scheduler reset) to ensure stability. We utilize the AdamW optimizer ($\beta_1{=}0.9, \beta_2{=}0.95$, weight decay $1{\times}10^{-4}$) with cosine scheduling and linear warm-up. We track an Exponential Moving Average (EMA, decay $0.9999$) with shadow weights on CPU for inference.

\paragraph{Stage 1: Modality Adaptation.}
Branches are trained independently under modality-specific flow-matching supervision to repurpose RGB priors for geometric and kinetic modeling without gradient interference. We use a learning rate of $2{\times}10^{-5}$ with $500$ warm-up steps and flow weight $\lambda_{\text{flow}}{=}0.5$.

\paragraph{Stage 2: Frozen-Backbone Joint Attention.}
We freeze the backbone and instantiate the 10 JA modules. To preserve established representations, we employ \textbf{zero-initialization} for the output projections of the JA modules. The learning rate is set to $5{\times}10^{-5}$ with $200$ warm-up steps. We apply Branch Dropout ($p_{\text{drop}}{=}0.2$) on $\{\text{depth}, \text{flow}\}$ to enhance cross-modal robustness.

\paragraph{Stage 3: Full-Parameter Joint SFT.}
We unfreeze the entire model for joint fine-tuning on the Rynn4DDataset 1.0.
We employ the learning rate as $1{\times}10^{-5}$ for all trainable parameters. Branch Dropout is reduced to $p_{\text{drop}}{=}0.1$.

\begin{table}[t]
\centering
\caption{Stage-wise training configuration. Effective batch size is per-GPU batch ($1$) $\times$ gradient accumulation (2-4) $\times$ $N_{\text{GPU}}$.}
\label{tab:training_hparams}
\setlength{\tabcolsep}{5pt}
\begin{tabular}{lccc}
\toprule
 & \textbf{Stage 1} & \textbf{Stage 2} & \textbf{Stage 3} \\
\midrule
\rowcolor{gray!10}Fusion mode & none & joint (frozen bb.) & joint (full SFT) \\
Trainable params & all branches & JA + mod.\ embed. & all parameters \\
\rowcolor{gray!10}Learning rate & $2{\times}10^{-5}$ & $5{\times}10^{-5}$ & $1{\times}10^{-5}$ \\
LR warm-up steps & $500$ & $200$ & $500$ \\
\rowcolor{gray!10}$\lambda_{\text{flow}}$ & $0.5$ & $1.0$ & $1.0$ \\
Branch Dropout & --- & $0.2$ & $0.1$ \\
\bottomrule
\end{tabular}
\end{table}

\paragraph{Resource and Optimization.}
All stages train at $81 \times 480 \times 640$ (yielding $T=21$ latent frames under the causal VAE's $4\times$ temporal compression, i.e., $T_{\text{latent}}=(T_{\text{pixel}}-1)/4+1$) with \texttt{bf16} mixed precision and gradient checkpointing. Stages 2 and 3 leverage DeepSpeed ZeRO-2 with optimizer offload to manage memory for additional JA parameters.

\begin{figure}[t]
  \centering
    \includegraphics[width=\linewidth]{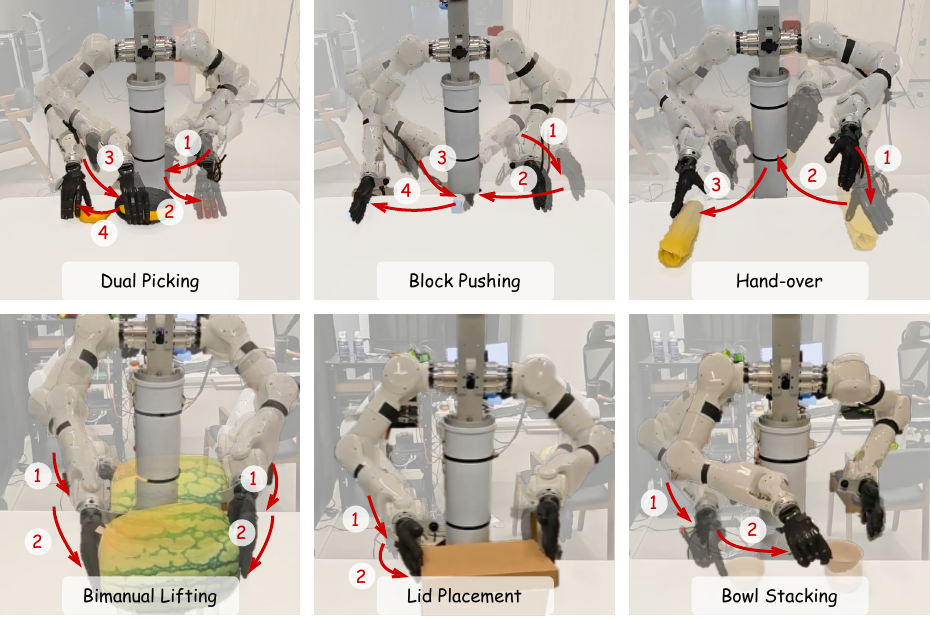}\\
    \caption{\textbf{Real-world Manipulation Benchmark.} We establish a comprehensive evaluation suite comprising six diverse tasks to assess the model's performance in open-world manipulation, providing a rigorous testbed for our 4D world model.}
  \label{fig:tasks}
\end{figure}

\subsubsection{\our-Policy}
We utilize \our\ as a frozen 4D vision encoder. The model operates at $480 \times 640$ resolution, producing $T=21$ latent frames (after VAE temporal compression with a ratio of 4). At each inference step, we perform single-step feature extraction at diffusion timestep $t=500$, capturing intermediate hidden states from block 15 of the transformer. By concatenating the 3072-dimensional features from the RGB, depth, and optical flow branches, we obtain a comprehensive 4D representation $F_p$.

At each decision step, we take a single RGB observation frame as input. The condition frame is center-cropped to $480 \times 640$ and normalized to $[-1, 1]$. The extracted spatiotemporal features are reshaped and fed into Flow Former. This compresses the high-dimensional \our features into a fixed-size representation suitable for policy decoding.

We employ a flow matching policy head with 4-step Euler ODE sampling at inference time. Despite the multi-step sampling in the action space, the policy maintains high efficiency because the heavy 4D backbone features are only computed once. The policy predicts action chunks of length 10, where each action is 54-dimensional. During deployment, the model executes 10 actions open-loop before re-querying the visual backbone, yielding an effective control frequency of $\sim$9 Hz.

We train with AdamW optimizer using a learning rate of $1 \times 10^{-4}$, $\beta = (0.9, 0.9)$, and weight decay 0.05. We employ a tri-stage learning rate schedule: 2\% linear warmup, 8\% constant hold, and 90\% cosine decay to $10^{-6}\times$ the peak learning rate. The \our backbone is kept frozen throughout training; only the Flow Former and flow matching policy head are optimized. Training uses mixed precision with a batch size of 1 per GPU for 100 epochs.

\subsection{Setups and Baselines}
\noindent
\textbf{World model metrics.}
To evaluate the generative performance and physical fidelity of our world model, we conduct benchmarks on a held-out test set of 50 video sequences, randomly sampled from~\citep{wu2024robomind,liu2024rdt,jiang2025galaxea}. We assess the model across three axes:
\begin{itemize}[labelsep=0.6em, leftmargin=1.2em,itemindent=0em]
    \item \textbf{(1) \textit{Visual Synthesis Quality}}: generative fidelity, following~\citep{zhou2025pai} (IQ, MS, SC, Subj.) and pixel-level alignment (PSNR, SSIM, LPIPS) with ground truth;

    \item \textbf{(2) \textit{Geometric Accuracy}}: evaluating structural integrity via depth estimation metrics (AbsRel, $\delta_1 < 1.25$);

    \item \textbf{(3) \textit{Temporal Motion Consistency}}: measuring dynamic precision through the optical-flow endpoint error (EPE/AEPE) in pixel-displacement space.
\end{itemize}
Although the flow branch is trained on color-coded flow representations, all quantitative motion metrics are computed in flow-vector space after decoding. Detailed metric definitions are provided in Appendix~\ref{sec:metric}, and implementation details of the compared 4D world models are in Appendix~\ref{sec:baseline_impl}.

\noindent
\textbf{Real-world task metric.}
The primary evaluation metric is the Success Rate, defined as the percentage of successful task completions over 35 consecutive real-world trials. A trial is considered successful if the robot completes the task within 120 seconds.

\noindent
\textbf{Policy baselines.}
We compare \our-Policy against two groups of baselines, all fine-tuned on the same real-robot demonstrations as ours: (1) policy-learning methods, including Diffusion Policy (DP)~\citep{chi2025diffusion}, the generalist foundation policies $\pi_0$~\citep{black2024pi_0} and $\pi_{0.5}$~\citep{intelligence2025pi_}, and the video-prediction-driven policies VPP~\citep{hu2024video} and Mimic-Video~\citep{pai2025mimic}; (2) generative world models repurposed for control, marked with $^*$ in Tab.~\ref{tab:real}: TesserAct$^*$~\citep{zhen2025learning}, a 4D world model that generates RGB, depth, and normal videos, and Wan-2.2-TI2V-5B$^*$~\citep{wang2025wan}, a video model that generates RGB. For these $^*$ variants, we extract their intermediate latent features and feed them into \our-Policy, so the comparison isolates the benefit of \our's internal 4D (RGB-DF) representation over the baselines' latents.

\noindent
\textbf{Hardware platform.}
For real-world data collection and policy evaluation, we utilize the TIANJI M6 robot equipped with a WUJI HAND. A RealSense D435i camera is integrated to capture first-person view (FPV) images. Please refer to our Appendix.~\ref{sec:body} for more details.

\noindent
\textbf{Robot tasks.}
\label{sec:tasks}
To demonstrate the generalization ability of our pipeline across diverse manipulation challenges, we evaluate our method on six distinct tasks that span varying levels of bimanual coordination, contact richness, and precision:
\begin{itemize}[labelsep=0.6em, leftmargin=1.2em,itemindent=0em]
    \item \textbf{(1) \textit{Dual Picking}}: The robot uses its left arm to pick an apple and its right arm to pick a banana from a plate, sequentially placing both objects onto the tabletop.

    \item \textbf{(2) \textit{Block Pushing}}: A sequential pushing task where the left arm pushes a large block from the left zone to the center, after which the right arm takes over to push it to the designated right target zone.

    \item \textbf{(3) \textit{Hand-over}}: An intra-robot transfer task where the left gripper picks up a cabbage and hands it over to the right gripper, which then completes the placement.

    \item \textbf{(4) \textit{Bimanual Lifting}}: A heavy-load coordination task where both arms must synchronously lift a watermelon plush from the table and accurately place it into a tray.

    \item \textbf{(5) \textit{Lid Placement}}: A precision-oriented task requiring the robot to pick up a lid and accurately align it to cover a cardboard box.

    \item \textbf{(6) \textit{Bowl Stacking}}: There are two small bowls on the table; the task involves picking up one bowl and carefully stacking it on top of the other.
\end{itemize}

As shown in Fig.~\ref{fig:tasks}, these tasks collectively challenge the model's proficiency in dual-arm synergy, temporal sequencing, and long-horizon interaction.
Tab.~\ref{tab:dataset_stats} summarizes the per-task demonstration
budget: the full corpus is used to train \our, exposing it to a rich
diversity of physical interactions, while a fixed budget of 200 episodes
per task is allocated for training \our-Policy.
For each task, we evaluate the model's generalization by applying significant randomization to the initial environment state. This includes varying the 6-DoF poses of task-relevant objects (\eg fruits) in terms of both workspace coordinates and axial rotations.

\begin{table}[t]
\centering
\caption{Statistics of the Real-World Dataset used for training.}
\label{tab:dataset_stats}
\begin{tabular}{lp{6cm}cc}
\hline
\textbf{Task Name} & \textbf{Description} & \textbf{RynnWorld-4D} & \textbf{RynnWorld-4D-Policy} \\ \hline
\rowcolor{gray!10}Dual Picking & Pick apple and banana from plate to table. & 500 & 200 \\
Block Pushing & Push block from left to center, then to the right. & 500 & 200 \\
\rowcolor{gray!10}Hand-over & Hand-over cabbage from left to right hand and place. & 300 & 200 \\
Bimanual Lifting & Lift a watermelon pillow and place it into a tray. & 500 & 200 \\
\rowcolor{gray!10}Lid Placement & Pick the lid and place it precisely on top of the box. & 300 & 200 \\
Bowl Stacking & Stack one bowl onto another. & 300 & 200 \\
\rowcolor{headerpurple!60} \textbf{Total} & & \textbf{2,400} & \textbf{1,200} \\ \hline
\end{tabular}
\end{table}

\begin{figure}
  \centering
    \includegraphics[width=\linewidth]{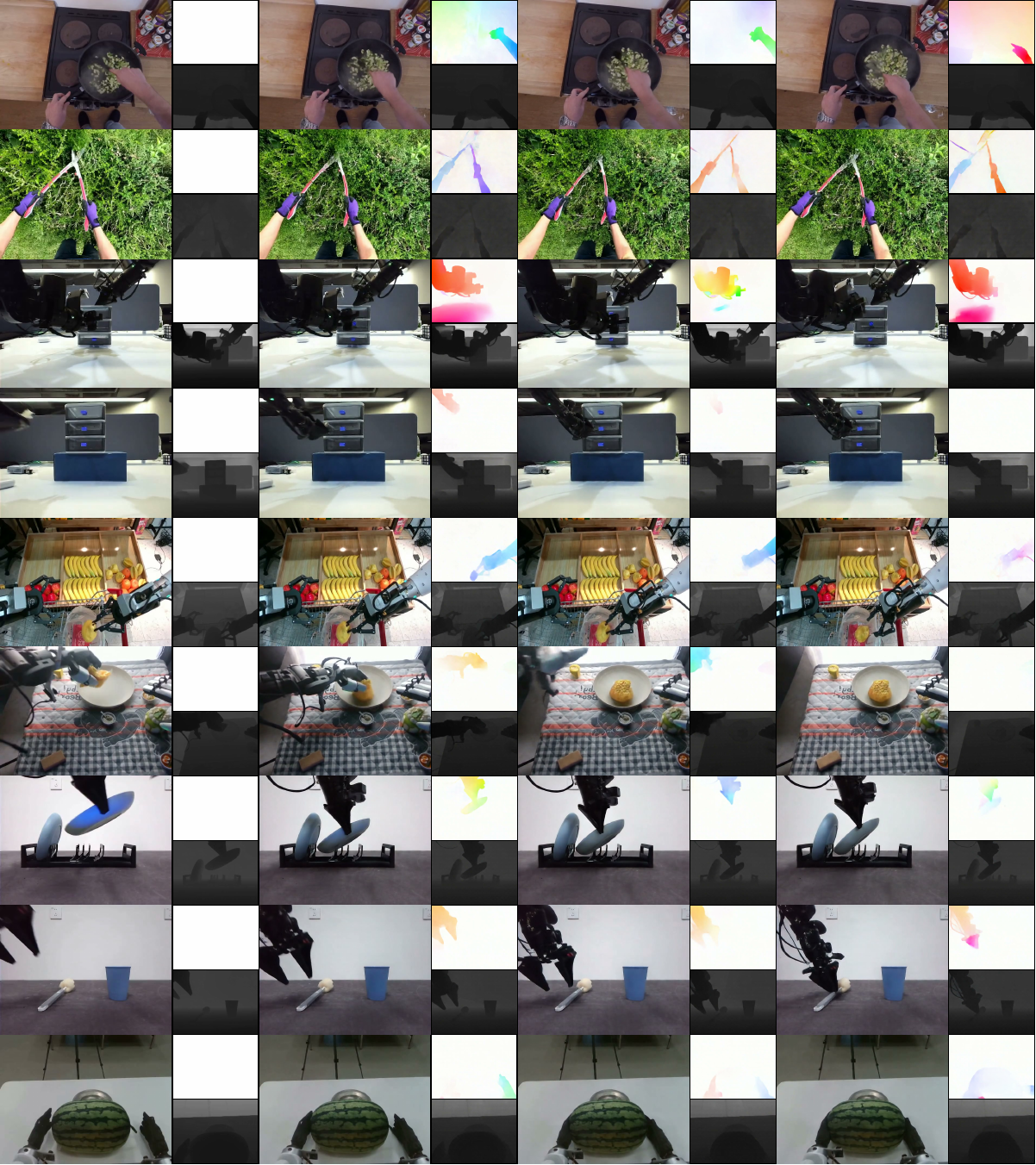}\\
  \caption{\textbf{Qualitative results of \our.} Starting from a single RGB-D image and a text prompt, \our synchronously generates RGB video, depth maps, and optical flow, preserving sharp geometric structures and producing temporally consistent motion fields.}
  \label{fig:vis-flowoworld}
\end{figure}

\subsection{Results}
\noindent
\textbf{4D World Modeling.}
As shown in Tab.~\ref{tab:exp}, compared to state-of-the-art video generation models such as Wan~\citep{wang2025wan} and CogVideoX~\citep{yang2024cogvideox}, \our maintains highly competitive visual quality (IQ) while significantly outperforming them in reconstruction fidelity. This indicates that while general video models excel at creative synthesis, our model is better at preserving the structural and textural integrity of the scene during evolution, benefiting from the mutual regularization between depth and flow branches. When compared to existing 4D world models, \our demonstrates a clear advantage in both spatial and temporal accuracy. In terms of geometry, our model achieves a $\delta_1$ of 0.610, nearly doubling the performance of 4DNeX~\citep{chen20254dnex} (0.327) and TesserAct~\citep{zhen2025learning} (0.279). Regarding motion, \our uniquely provides synchronized optical flow with a low AEPE of 0.170, whereas most baseline 4D models lack the capability to produce explicit motion fields.

\begin{figure}
  \centering
    \includegraphics[width=0.9\linewidth]{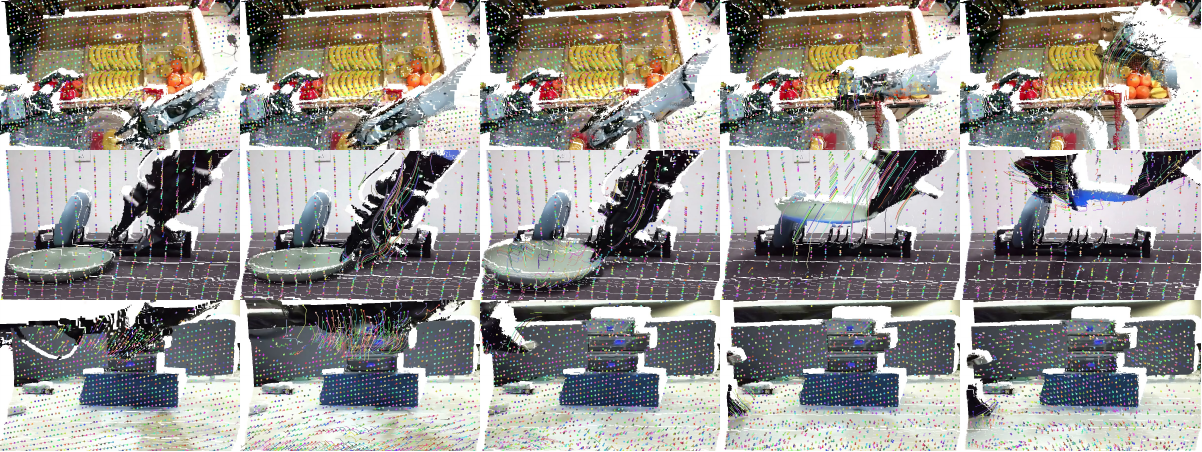}\\
  \caption{\textbf{Visualization of 3D scene flow derived from \our's generations.} By back-projecting the co-generated depth and optical flow into camera space (Sec.~\ref{sec:3d_reconstruction}), \our lifts its projective 4D predictions into explicit 3D scene flow. The trajectories, rendered as colored trails over the depth-ordered point cloud, are temporally coherent and geometrically plausible, confirming that the generated modalities are geometrically consistent with each other rather than mere visual hallucinations.}
  \label{fig:3dflow}
\end{figure}

As visualized in Fig.~\ref{fig:vis-flowoworld}, the generated depth and flow maps are not only internally consistent but also precisely aligned with the RGB texture changes. This result validates our core hypothesis: jointly modeling RGB, depth, and flow within a single diffusion loop acts as a powerful geometric and motion regularizer. Furthermore, Fig.~\ref{fig:3dflow} presents the 3D scene flow lifted from the co-generated depth and optical flow (Sec.~\ref{sec:3d_reconstruction}), demonstrating that our projective 4D predictions can be faithfully unprojected into coherent 3D motion trajectories. More qualitative results are provided in Appendix~\ref{appendix:rynnworld_4d}.

\begin{table*}[t]
 \centering
    \caption{\textbf{Quantitative evaluation of 4D world modeling quality.} Metrics span visual synthesis (RGB), geometric structure (Depth), and temporal dynamics (Optical Flow). N/A denotes that the baseline lacks the native capability to generate specific modalities.}
    \label{tab:exp}
    \small
    \setlength{\tabcolsep}{1.3pt}
    \begin{tabular}{l ccc ccc c cc c}
    \toprule
    \multirow{2}{*}{\textbf{Method}} & \multicolumn{7}{c}{\textbf{RGB (Generation \& Fidelity)}} & \multicolumn{2}{c}{\textbf{Depth}} & \multicolumn{1}{c}{\textbf{Flow}} \\
    \cmidrule(lr){2-8} \cmidrule(lr){9-10} \cmidrule(lr){11-11}
    & IQ $\uparrow$ & MS $\uparrow$ & SC $\uparrow$ & Subj. $\uparrow$ & SSIM $\uparrow$ & PSNR $\uparrow$ & LPIPS $\downarrow$ & AbsRel $\downarrow$ & $\delta_1 \uparrow$ & AEPE $\downarrow$  \\
    \midrule
    \multicolumn{11}{c}{\textit{Video Generation Models}} \\
    \rowcolor{gray!10} CogVideoX~\citep{yang2024cogvideox} & 0.604 & 0.976 & 0.866 & 0.917 & 0.534 & 12.17 & 0.577 & N/A & N/A & N/A \\
    Wan-2.2-TI2V-5B~\citep{wang2025wan} & 0.555 & 0.970 & 0.886 & 0.909 & 0.593 & 14.54 & 0.489 & N/A & N/A & N/A \\
    \rowcolor{gray!10} Wan-2.1-I2V-14B~\citep{wang2025wan} & \textbf{0.684} & 0.988 & 0.891 & 0.956 & 0.536 & 12.72 & 0.568  & N/A & N/A & N/A \\

    \midrule
    \multicolumn{11}{c}{\textit{4D World Models}} \\
    \rowcolor{gray!10} Free4D~\citep{liu2025free4d} & 0.354 & 0.993 & 0.787 & 0.848 & 0.492 & 12.40 & 0.597 & 0.804 & 0.179 & N/A \\
    TesserAct~\citep{zhen2025learning} & 0.608 & 0.992 & 0.904 & 0.956 & 0.693 & 16.91 & 0.335 & 0.699 & 0.279 & N/A \\
    \rowcolor{gray!10} 4DNeX~\citep{chen20254dnex} & 0.637 & 0.994 & 0.917 & 0.986 & 0.649 & 14.47 & 0.404 & 0.423 & 0.327 & N/A \\
    \rowcolor{headerpurple!60} \textbf{\our}  & 0.635 & \textbf{0.995} & \textbf{0.957} & \textbf{0.992} & \textbf{0.754} & \textbf{17.85} & \textbf{0.269} & \textbf{0.310} & \textbf{0.610} & \textbf{0.170} \\

    \midrule
    \multicolumn{11}{c}{\textit{Ablation Study}} \\
    \rowcolor{gray!10} Independent Branches  & 0.613 & 0.986 & 0.922 & 0.971 & 0.683 & 17.26 & 0.346 & 0.737 & 0.245 & 0.247 \\
    w/o Modality Adaptation & 0.621 & 0.992 & 0.952 & 0.975 & 0.699 & 17.85 & 0.303 & 0.507 & 0.479 & 0.231 \\
    \rowcolor{gray!10} w/o 4D Pre-training & 0.615 & 0.982 & 0.879 & 0.938 & 0.651 & 16.25 & 0.344 & 0.797 & 0.263 & 0.729 \\
    w/o RoPE in JA & 0.628 & 0.990 & 0.935 & 0.980 & 0.710 & 17.10 & 0.295 & 0.420 & 0.450 & 0.210 \\
    \rowcolor{gray!10} Shared FFN & 0.618 & 0.985 & 0.902 & 0.965 & 0.695 & 16.50 & 0.320 & 0.580 & 0.380 & 0.280 \\

    \bottomrule
    \end{tabular}
\end{table*}

\noindent
\textbf{Policy Learning.}
Tab.~\ref{tab:real} summarizes the performance of \our-Policy across various robotic manipulation tasks, where it consistently outperforms state-of-the-art baselines spanning imitation learning (DP~\citep{chi2025diffusion}), generalist foundation policies ($\pi_0$~\citep{black2024pi_0}, $\pi_{0.5}$~\citep{intelligence2025pi_}), video-prediction-driven policies (VPP~\citep{hu2024video}, Mimic-Video~\citep{pai2025mimic}), and generative world models repurposed for control (TesserAct$^*$~\citep{zhen2025learning}, Wan-2.2-TI2V-5B$^*$~\citep{wang2025wan}).

Notably, in tasks requiring high spatial precision such as \textit{Lid Placement} and \textit{Bowl Stacking}, \our-Policy achieves success rates of 65.71\%, surpassing the next best baseline (DP) by 8.57\%.
Even more striking is the \textit{Hand-over} task, a challenge involving dynamic object transfer where foundation models struggle.
This performance gap stems from two fundamental limitations of current foundation models: first, their pre-training data is predominantly biased towards parallel-jaw grippers, lacking the inherent priors for the complex dexterous hand coordination. Second, in a hand-over scenario, 2D-based models struggle to reason about the relative 3D distance and potential self-occlusion between two high-DOF end-effectors.
Furthermore, while these 2D policies must implicitly recover complex transfer dynamics from appearance residuals in the RGB stream, \our-Policy benefits from explicit kinetic and geometric cues provided by the world model's internal 4D latents, proving that predictive 4D representations are essential for tasks requiring precise temporal and spatial coordination where pure generative 2D priors fall short.

\begin{table*}[t]
 \centering
    \caption{\textbf{Success rates (\%) on robotic manipulation tasks.} We compare \our-Policy against state-of-the-art policy learning baselines and foundation models across six challenging tasks. $^*$ denotes generative world models fine-tuned on our real-robot data; their intermediate latent features are fed into the \emph{same} Flow Former for action decoding.}
    \label{tab:real}
    \small
    \setlength{\tabcolsep}{4pt}
 \begin{tabular}{l *{6}{>{\centering\arraybackslash}m{1.4cm}}}
 \toprule
  \textbf{Method}
 & \textbf{Dual Picking}
 & \textbf{Block Pushing}
 & \textbf{Hand-over}
 & \textbf{Bimanual Lifting}
 & \textbf{Lid Placement}
 & \textbf{Bowl Stacking} \\
 \midrule
 \rowcolor{gray!10} DP~\citep{chi2025diffusion} & 77.14 & 85.71 & 17.14 & 88.57 & 57.14 & 57.14 \\
 $\pi_0$~\citep{black2024pi_0} & 88.57 & 94.29 & 2.86 & 91.43 & 34.29 & 51.43 \\
 \rowcolor{gray!10} $\pi_{0.5}$~\citep{intelligence2025pi_} & \textbf{94.29} & \textbf{100.00} & 0.00 & 94.29 & 37.14 & 42.86 \\
 VPP~\citep{hu2024video}  & 85.71 & 91.43 & 5.71 & 94.29 & 57.14 & 51.43 \\
 \rowcolor{gray!10} Mimic-Video~\citep{pai2025mimic} & 80.00 & 88.57 & 8.57 & 91.43 & 51.43 & 54.29 \\
 TesserAct$^*$~\citep{zhen2025learning} & 91.43 & 94.29 & 17.14 & 94.29 & 57.14 & 60.00 \\
 \rowcolor{gray!10} Wan-2.2-TI2V-5B$^*$~\citep{wang2025wan} & 77.14 & 91.43 & 14.29 & 91.43 & 57.14 & 60.00 \\
 \rowcolor{headerpurple!60} \textbf{\our-Policy} & \textbf{94.29} & 97.14 & \textbf{28.57} & \textbf{97.14} & \textbf{65.71} & \textbf{65.71} \\

 \midrule[\heavyrulewidth]
 \multicolumn{7}{c}{\textit{Ablation Study}} \\
 \rowcolor{gray!10} ResNet~\citep{he2016deep} as encoder  & 71.43 & 88.57 & 11.43 & 85.71 & 51.43 & 60.00 \\
 RGB  & 77.14 & 91.43 & 14.29 & 91.43 & 57.14 & 60.00 \\
 \rowcolor{gray!10} RGB + Depth  & 91.43 & 91.43 & 28.57 & 97.14 & 60.00 & 62.86 \\
 RGB + Optical Flow & 85.71 & 88.57 & 20.00 & 88.57 & 54.29 & 62.86 \\
 \bottomrule
 \end{tabular}
\end{table*}

\subsection{Ablation Study}
We conduct extensive ablation studies to validate our architectural design and training strategies.

\noindent
\textbf{Predictive Representations for Manipulation.}
The starred rows in Tab.~\ref{tab:real} form a controlled comparison that fixes our Flow Former and action head and varies only \emph{what the world model predicts}. Wan-2.2-TI2V-5B$^*$~\citep{wang2025wan} forecasts RGB only, forcing the decoder to infer structure and motion from pixel changes; together with VPP~\citep{hu2024video} and Mimic-Video~\citep{pai2025mimic}, these RGB-only predictors trail \our-Policy by 10.79\% on average, with the largest gap on the motion-dominated \textit{Hand-over}. TesserAct$^*$ additionally predicts depth and surface normals, which improves over the RGB-only predictors on most tasks; yet normals encode only \emph{static} orientation and carry no inter-frame correspondence, so on the precision-critical \textit{Lid Placement} and \textit{Bowl Stacking} it closes none of the gap (57.14\% and 60.00\%, matching the strongest RGB-only predictor), and on the motion-dominated \textit{Hand-over} it remains 11.43\% behind ours. Optical flow supplies exactly this missing correspondence: paired with depth it back-projects into 3D scene flow (Sec.~\ref{sec:3d_reconstruction}), a per-point displacement field of the same nature as the end-effector motion the policy must output. RGB-DF is thus the representation closest to the action space, and because every variant here reads its internal latents through the same decoder, the observed gaps reflect the quality of the representation itself rather than any decoding difference.

\noindent
\textbf{Effectiveness of Tri-branch Fusion.}
To verify the necessity of synchronized generation, we compare \our with the \textbf{Independent Branches} baseline, where three diffusion branches are trained separately. As shown in Tab.~\ref{tab:exp}, while independent branches can generate individual modalities, they suffer from significant performance degradation in depth (0.737 vs. 0.310 AbsRel) and flow (0.247 vs. 0.170 AEPE). This confirms that our mutual feature interaction mechanism is crucial for enforcing cross-modal consistency and physical fidelity in 4D sequences.

\noindent
\textbf{Necessity of Modality Adaptation.}
The comparison between the full model and the \textbf{w/o Modality Adaptation} baseline (which skips the initial modality-specific adaptation and proceeds directly to joint tri-branch training) highlights the efficacy of our phased training strategy. Without this first stage, the depth and flow branches struggle to bridge the gap between their inherited RGB priors and their specific modality distributions (Tab.~\ref{tab:exp}). This leads to a substantial drop in geometric accuracy ($\delta_1$ decreases from 0.610 to 0.479) and compromised motion precision, proving that modality-specific adaptation is a prerequisite for effective multi-modal fusion.

\noindent
\textbf{Significance of Large-scale 4D Pre-training.}
We evaluate the necessity of Rynn4DDataset 1.0 by comparing our full model with the \textbf{w/o 4D Pre-training} variant, which is trained exclusively on a limited set of task-specific robotic manipulation data. Omitting the large-scale pre-training on Rynn4DDataset 1.0 leads to a severe performance collapse across all dimensions, with the AEPE surging from 0.170 to 0.729 (Tab.~\ref{tab:exp}). These results underscore that task-specific data alone lacks the diversity required to master complex spatio-temporal dynamics.

\noindent
\textbf{Effectiveness of Predictive Latents.}
To verify the importance of the \our latent space, we compare our model against a baseline that replaces the predictive encoder with a standard ResNet-18~\citep{he2016deep} image encoder (\textbf{ResNet~\citep{he2016deep} as encoder} in Tab.~\ref{tab:real}). The significant performance drop across all tasks—most notably in \textit{Dual Picking} where success falls from 94.29\% to 71.43\%—highlights that static 2D features are insufficient for complex tasks. This confirms that the temporal and spatial dynamics captured in \our's predictive representations are crucial for robust policy learning.

\noindent
\textbf{Impact of 4D Modalities.}
We further analyze the individual contribution of each modality (Tab.~\ref{tab:real}). Using only RGB latents (\textbf{RGB}) yields sub-optimal success rates as the policy lacks explicit structural grounding. Integrating depth (\textbf{RGB + Depth}) provides significant gains in tasks requiring spatial precision, such as \textit{Hand-over} and \textit{Bimanual Lifting}. Meanwhile, adding optical flow (\textbf{RGB + Optical Flow}) enhances motion-sensitive tasks. The full \our-Policy, combining all three, achieves the best performance, confirming that the synergy of visual context, spatial geometry, and kinetic dynamics is essential for robust robotic manipulation.

\noindent
\textbf{Role of 3D RoPE in Joint Attention.}
We disable the 3D Rotary Positional Embedding (RoPE) inside the Joint Cross-Modal Attention modules (\textbf{w/o RoPE in JA} in Tab.~\ref{tab:exp}) to decouple spatial coordinates from cross-modal interactions. Although intra-modal self-attention preserves local spatial structure, removing the shared positional bias in the cross-branch pathway disrupts the geometric correspondence between modality-specific features at identical $(u,v)$ coordinates. This misalignment manifests as a significant decay in reconstruction fidelity, with $\delta_1$ dropping from 0.610 to 0.450 and AEPE rising from 0.170 to 0.210. These results highlight that 3D RoPE serves as a critical alignment bridge, enabling the JA modules to achieve spatially-aware feature fusion at the pixel level rather than mere global semantic averaging.

\noindent
\textbf{Per-branch FFN vs.\ Shared FFN.}
By default, our architecture employs independent feed-forward networks (FFNs) for each branch, initialized from the pre-trained RGB backbone to preserve specialized representation manifolds. Replacing these with a single FFN shared across all modalities (\textbf{Shared FFN} in Tab.~\ref{tab:exp}) leads to a systemic performance collapse. This suggests that the latent spaces of RGB textures, depth manifolds, and motion fields are inherently heterogeneous; forcing them through a shared non-linear transformation induces catastrophic interference in feature representation. The resulting decline in geometric accuracy (AbsRel 0.580 / $\delta_1$ 0.380) and motion stability (AEPE 0.280) empirically validates that modality-specific FFNs are essential to mitigate cross-modal distribution shifts and maintain high-fidelity 4D generation. Although this shared variant reduces parameter overhead, the substantial drop in generative fidelity underscores the necessity of modality-specific capacity in 4D world modeling.

%% file: Sec/5-conclusion.tex
\section{Conclusion}
In this paper, we presented \textbf{\our}, a novel framework that shifts the paradigm of generative world modeling from 2D pixel sequences to consistent 4D scene evolution. By introducing a lightweight yet expressive representation consisting of synchronized RGB, depth, and optical flow (RGB-DF), we effectively bridge the gap between the scalability of video diffusion models and the geometric rigor required for robotic manipulation. Our core contributions include the development of a tri-branch architecture that ensures cross-modal consistency through mutual feature interactions, and the curation of \textbf{Rynn4DDataset 1.0}, the large-scale hybrid dataset specifically designed for training 4D generative models. Furthermore, we demonstrated that \textbf{\our-Policy} can effectively leverage these predictive 4D representations as an implicit world model, enabling high-frequency, closed-loop robotic control that outperforms existing 2D-based baselines. Extensive experiments show that our approach not only generates high-fidelity 4D futures but also significantly enhances the success rate and precision of downstream manipulation tasks. We believe \our provides a promising foundation for building general-purpose embodied intelligence capable of understanding and interacting with the complex 3D world.

\noindent
\textbf{Limitation.}
Despite the reactive capabilities of \our-Policy, our framework has several limitations. First, the 4D sequence generation relies on a diffusion denoising process, which introduces computational overhead. Currently, our implementation achieves an effective control frequency of approximately 9 Hz on an NVIDIA RTX 5090 GPU; while sufficient for many tasks, this latency remains a bottleneck for ultra-high-frequency control. Second, our model is primarily optimized for egocentric perspectives. Extending 4D spatio-temporal consistency to multi-view systems or collaborative multi-robot setups remains an open challenge for future research.

%% file: Sec/Appendix.tex
\section{Real Robot System Setup}
\label{sec:body}
\begin{figure}
  \centering
    \includegraphics[width=\linewidth]{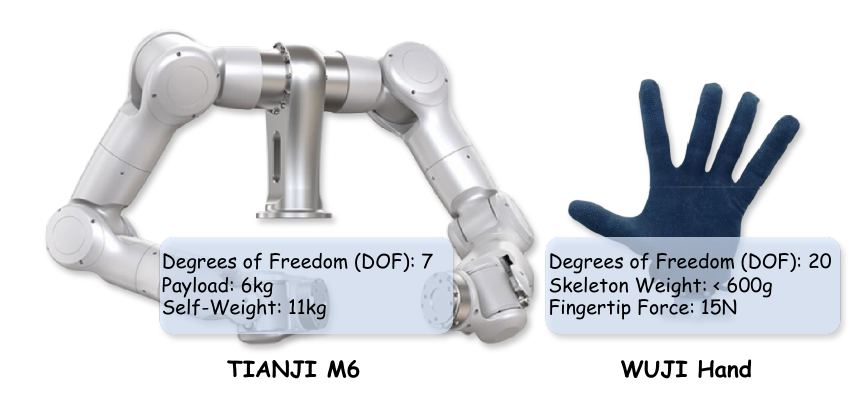}\\
  \caption{\textbf{Standardized Experimental Platform.} All real-world experiments are conducted on a unified hardware configuration consisting of the TIANJI M6 7-DOF robotic arm and the 20-DOF WUJI dexterous hand. This integrated system provides the high-precision control and high-dimensional actuation space required for complex manipulation tasks.}
  \label{fig:body}
\end{figure}

Our real robot is built on the TIANJI M6 and WUJI Hand, as shown in Fig.~\ref{fig:body}.
The policy’s inference frequency is set at 50 Hz. The commands are sent with a delay kept between 18 and 30 milliseconds. The low-level interface operates at a frequency of 500 Hz, ensuring smooth real-time control. The communication between the control policy and the low-level interface is realized through LCM (Lightweight Communications and Marshalling).

\begin{figure}
  \centering
    \includegraphics[width=0.45\linewidth]{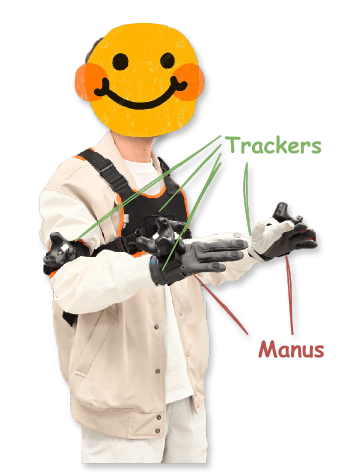}\\
  \caption{Operator to collect real world data.}
  \label{fig:operator}
\end{figure}

We collect real-world demonstration data through a teleoperation system, as shown in Fig.~\ref{fig:operator}. Our hardware setup consists of dual Tianji 7-DOF robotic arms and dual Wuji 20-DOF dexterous hands, yielding a total of 54 degrees of freedom.

For arm control, the human operator wears five HTC Vive trackers (mounted on the chest, both wrists, and both upper arms). The system computes wrist-to-chest relative transforms at 100-120 Hz and feeds them into a Pinocchio-based inverse kinematics solver running in a separate process. The resulting joint commands are further smoothed by a Ruckig trajectory generator with velocity, acceleration, and jerk constraints before being sent to the robot arms at 200 Hz.

For hand control, the operator wears Manus data gloves. The raw glove signals are converted to a 21-point MediaPipe hand skeleton format and retargeted to the 20-DOF Wuji hand joint space via a dedicated retargeting module, with an exponential moving average filter applied for motion smoothing.

\section{Additional Details on Evaluation Metrics}
\label{sec:metric}
This section provides technical definitions and implementation details for all evaluation metrics used in our experiments.

\subsection{Generative Quality}
We adopt a subset of task-aligned metrics from PAI-Bench~\citep{zhou2025pai} to assess the visual and temporal quality of generated videos:
\begin{itemize}[labelsep=0.6em, leftmargin=1.2em,itemindent=0em]
\item \textbf{Imaging Quality (IQ):} Evaluates low-level visual fidelity including clarity, noise levels, and compression artifacts. We employ the MUSIQ~\citep{ke2021musiq} predictor trained on the SPAQ dataset to compute a frame-level perceptual quality score.
\item \textbf{Motion Smoothness (MS):} Quantifies temporal coherence and physical plausibility of motion dynamics. It is computed as the reconstruction error between generated frames and those synthesized via the AMT-S~\citep{li2023amt} frame interpolation model, where lower interpolation error indicates smoother motion.
\item \textbf{Subject Consistency (SC):} Measures the identity stability of the primary subject across the video sequence. We compute the mean pairwise cosine similarity of DINO~\citep{zhang2022dino} ViT-B/16 features between the first frame and all subsequent frames.
\item \textbf{I2V-Subject (Subj.):} For image-to-video generation, this metric evaluates how faithfully the model preserves the identity of the input reference image throughout the generated sequence. It is computed as the DINO feature similarity between the conditioning image and each generated frame.
\item \textbf{PSNR (Peak Signal-to-Noise Ratio):} Measures pixel-wise reconstruction accuracy between generated and GT frames. Higher values indicate lower distortion. We report the mean PSNR across all frames (excluding the shared first frame) and all video samples.
\item \textbf{SSIM (Structural Similarity Index):} Evaluates the preservation of structural information by jointly considering luminance, contrast, and structural components between generated and GT frame pairs.
\item \textbf{LPIPS (Learned Perceptual Image Patch Similarity):} Measures perceptual distance between generated and GT frames using deep features from a pre-trained AlexNet. Lower values indicate higher perceptual similarity, and LPIPS is generally better aligned with human visual judgments than pixel-level metrics.
\end{itemize}

\subsection{Geometric Accuracy}
For methods that jointly predict depth maps, we evaluate structural fidelity of the estimated geometry using scale-invariant depth metrics:
\begin{itemize}[labelsep=0.6em, leftmargin=1.2em,itemindent=0em]
\item \textbf{Absolute Relative Error (AbsRel $\downarrow$):} Measures the mean relative deviation between the predicted depth $\hat{d}$ and ground-truth depth $d^*$. Since different methods may produce depth in arbitrary scales, we first perform scale-invariant alignment via median scaling: $s = \text{median}(d^*) / \text{median}(\hat{d})$, then compute:
\begin{equation}
\text{AbsRel} = \frac{1}{|\mathcal{V}|} \sum_{p \in \mathcal{V}} \frac{|s \cdot \hat{d}_p - d^*_p|}{d^*_p},
\end{equation}
where $\mathcal{V}$ denotes the set of valid pixels ($d^* > 0$).
\item \textbf{Threshold Accuracy ($\boldsymbol{\delta_1 < 1.25}$ $\uparrow$):} Reports the percentage of pixels whose depth ratio falls within a tolerance threshold:
\begin{equation}
\delta_1 = \frac{1}{|\mathcal{V}|} \sum_{p \in \mathcal{V}} \mathbb{1}\left[\max\left(\frac{s \cdot \hat{d}_p}{d^*_p},\; \frac{d^*_p}{s \cdot \hat{d}_p}\right) < 1.25\right].
\end{equation}
Higher values indicate better geometric alignment with the ground truth.
\end{itemize}

\subsection{Motion Consistency}
\label{app:metrics-flow}
For methods that predict optical flow, we evaluate the temporal dynamics accuracy:
\begin{itemize}[labelsep=0.6em, leftmargin=1.2em,itemindent=0em]
\item \textbf{Average End-Point Error (AEPE $\downarrow$):} Measures the accuracy of predicted optical flow fields against ground-truth flow. For each pixel $p$ with predicted flow $\hat{\mathbf{f}}_p = (\hat{u}_p, \hat{v}_p)$ and ground-truth flow $\mathbf{f}^*_p = (u^*_p, v^*_p)$, the end-point error is defined as:
\begin{equation}
\text{AEPE} = \frac{1}{|\mathcal{V}|} \sum_{p \in \mathcal{V}} \left\| \hat{\mathbf{f}}_p - \mathbf{f}^*_p \right\|_2 = \frac{1}{|\mathcal{V}|} \sum_{p \in \mathcal{V}} \sqrt{(\hat{u}_p - u^*_p)^2 + (\hat{v}_p - v^*_p)^2}.
\end{equation}
Although the flow branch operates on color-coded flow representations during training (to reuse the pretrained RGB video VAE), all quantitative flow metrics are computed in flow-vector space. Specifically, generated color-coded flow videos are first decoded back to flow vectors using the inverse color-wheel mapping before computing AEPE. The metric is averaged over all valid frames (excluding the first frame, which has no temporal reference) and all video samples. Lower AEPE indicates more accurate motion prediction.
\end{itemize}

\section{Preliminaries}
\label{sec:method:preliminaries}
Our study builds upon the Wan family of video generation models~\citep{wang2025wan}, a latent video diffusion transformer capable of generating temporally coherent video from a single input image or text prompt. The model consists of a 3D variational autoencoder $(\mathcal{E}, \mathcal{D})$ and a transformer-based diffusion model parameterized by $\Theta$. Given an input latent $z_0 = \mathcal{E}(V_0)$, the forward process follows the rectified flow formulation~\citep{esser2024scaling}, where the noised latent is generated by linear interpolation:
\begin{equation}
z_t = (1 - t)\, z_0 + t\,\epsilon, \quad \epsilon \sim \mathcal{N}(0, I)
\label{eq:zt_interp}
\end{equation}
with timestep $t \in [0, 1]$. The denoising process learns a velocity field $v_\Theta(z_t, t)$ that guides the transformation of noise back to data. The model is trained using a conditional flow matching~\citep{lipman2022flow}, with objective:
\begin{equation}
\mathcal{L}_{\text{CFM}} =
\mathbb{E}_{t, z_0, \epsilon}\!\left[\left\lVert
v_\Theta(z_t, t) - u_t(z_0 \mid \epsilon)
\right\rVert_2^2\right]
\label{eq:l_cfm}
\end{equation}
where $u_t$ is the target velocity derived analytically from the forward process. At inference, a sequence of latent frames is recovered by integrating $v_\Theta$ over time.

In the image-to-video (I2V) setting, the model is conditioned on an initial image $I_0$ encoded as $z_{\text{img}} = \mathcal{E}(I_0)$. The transformer-based denoiser $\mathcal{F}_\Theta$ autoregressively predicts video latents $\{z^{(f)}\}_{f=1}^F$, starting from $z_{\text{img}}$ and producing temporally consistent sequences. In the text-to-video (T2V) setting, the model is instead conditioned on a text prompt $p$ encoded as $z_{\text{text}}=\mathcal{T}(p)$ and starts the autoregressive generation from noise. The final video is reconstructed as $\hat{V} = \mathcal{D}(z^{(1)}, \dots, z^{(F)})$.

\section{Baseline Implementation Details}
\label{sec:baseline_impl}

\noindent
\textbf{Free4D.}
Free4D~\citep{liu2025free4d} lifts a single image into a dynamic 4D Gaussian Splatting (4DGS) representation. Given the first frame of each ground-truth video as input, we run the full Free4D pipeline: (1) the built-in ViewCrafter module with DUSt3R-based monocular depth estimation synthesizes 25 novel views from the input image via a video diffusion model at $576 \times 1024$ resolution; (2) COLMAP sparse reconstruction estimates camera poses and produces a sparse point cloud for Gaussian initialization; (3) the 4DGS model with HexPlane-based deformation fields is optimized for 10{,}000 iterations (3{,}000 static initialization + 7{,}000 joint optimization), with temporal resolution $[64, 64, 64, 150]$; (4) RGB and depth videos are rendered from the original camera viewpoint across all timesteps. Since the rendered depth is in arbitrary scale, we apply per-frame median scaling alignment before computing depth metrics.

\noindent
\textbf{4DNeX.}
4DNeX~\citep{chen20254dnex} is a feed-forward 4D scene generation framework that repurposes the Wan2.1-I2V-14B~\citep{wang2025wan}, fine-tuned with learnable domain embeddings and LoRA adapters, to jointly produce RGB appearance and per-pixel XYZ point-cloud geometry from a single image and a text prompt. We adopt the official variant, using the provided 4dnex-lora weights (rank $64$, fused at scale $0.5$).
For each sequence we extract the first frame of the ground-truth video as the conditioning image and use the corresponding caption, appending the official \texttt{POINTMAP\_STYLE.} suffix as required by the released model. For each sample we run $50$ denoising steps with classifier-free guidance scale $5.0$ and seed $42$, generating $49$ frames at $24$~fps. We obtain the depth video by taking the $z$-channel of the predicted pointmap, min--max normalized per sequence to $[0, 255]$.

\noindent
\textbf{TesserAct.}
TesserAct~\citep{zhen2025learning} is built upon CogVideoX-5b-I2V~\citep{yang2024cogvideox} and fine-tuned to jointly generate RGB, depth, and surface normal videos from a single initial frame and a language instruction. We adopt the official  checkpoint. Since TesserAct requires depth and normal maps as additional conditioning inputs, we first extract the initial RGB frame from each ground-truth video, then apply network to estimate monocular depth and surface normals. The three modalities are concatenated along the channel dimension to form a 9-channel input. We generate 49 frames at $640\times480$ resolution using 50 DDPM denoising steps with guidance scale 7.5, image guidance scale 1.5. The model outputs RGB, depth, and normal videos concatenated along the width axis; we split along width to obtain separate RGB and depth predictions for evaluation.

\section{Additional Qualitative Visualizations via RynnWorld-4D}
\label{appendix:rynnworld_4d}
\begin{figure}
    \centering
    \includegraphics[width=1.0\textwidth]{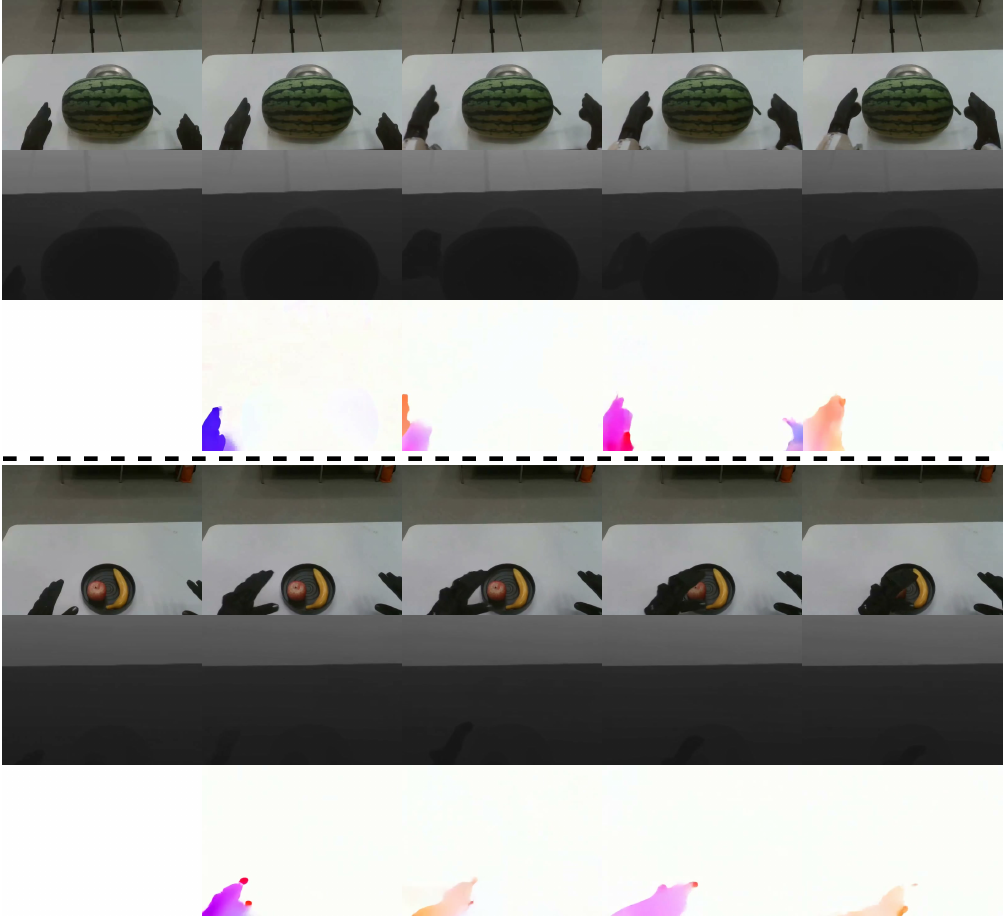}
    \caption{Extended qualitative results. Each row displays the generated RGB, depth, and optical flow sequences. The results highlight \our's ability to produce spatially and temporally coherent 4D predictions across various manipulation scenarios.}
    \label{fig:rynnworld_4d_vis}
\end{figure}

To further showcase the generation quality of \our, we provide extended paired visualizations of \textbf{RGB}, \textbf{Depth}, and \textbf{Optical Flow} generated via RynnWorld-4D (Fig.~\ref{fig:rynnworld_4d_vis}). \our synchronously predicts future RGB, depth maps, and optical flow from a single RGB-D observation. These results demonstrate:

(i) \textbf{Cross-modal Consistency:} The geometric structures in depth maps and motion boundaries in optical flow are precisely aligned with the RGB textures.

(ii) \textbf{Physical Fidelity:} The model accurately captures complex 4D dynamics, such as object displacements and multi-contact interactions, in both human-centric and robot-specific environments.

(iii) \textbf{Temporal Coherence:} The generated sequences maintain stability over time without significant flickering or structural morphing.

\begin{figure}[t]
  \centering
    \includegraphics[width=\linewidth]{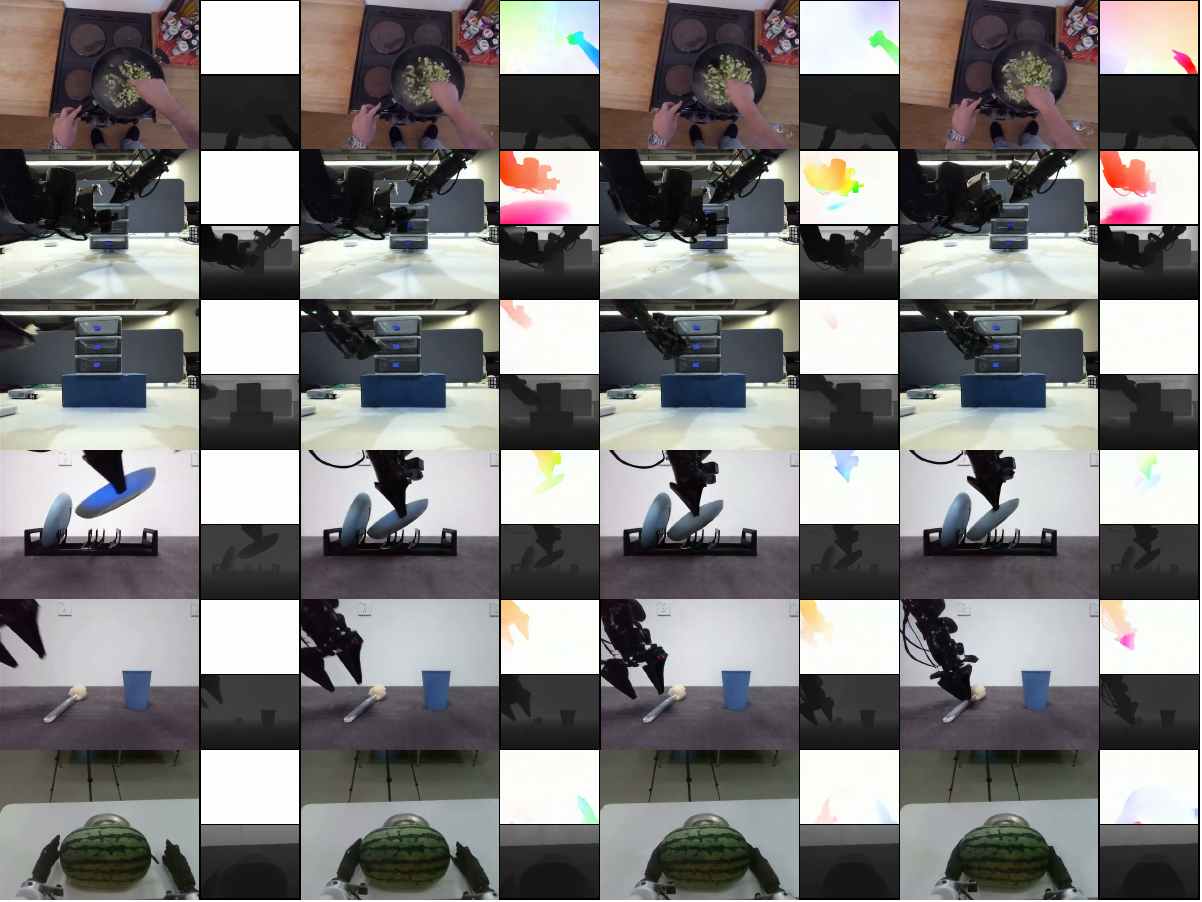}\\
  \caption{\textbf{Qualitative 4D predictions on our real-robot tasks.} Given a single RGB-D observation and the task instruction, \our generates synchronized RGB, depth, and optical-flow sequences on the six evaluated manipulation tasks, providing the predictive 4D representations consumed by \our-Policy.}
  \label{fig:vis-app}
\end{figure}

\noindent
In addition, Fig.~\ref{fig:vis-app} presents \our's synchronized RGB, depth, and optical-flow predictions on the six real-robot manipulation tasks, providing the predictive 4D representations consumed by \our-Policy.

\section{Representation Probing}
To examine whether RGB, depth, and optical flow (RGB-DF) is closer to the robot action space than RGB or RGB-D, we conduct a representation probing experiment following the low-level control regression protocol of LARYBench~\cite{nie2026lary}, as shown in Tab.~\ref{tab:representation_probe}, on its four regression datasets---CALVIN, VLABench, RoboCOIN, and AgiBot-Beta---which span simulated and real-robot manipulation as well as single-arm and bimanual embodiments. Each dataset uses the benchmark's native temporal stride ($5$ for CALVIN and VLABench, $10$ for RoboCOIN, $45$ for AgiBot-Beta): a sample is a frame pair $(\mathbf{I}_t, \mathbf{I}_{t+s})$ together with the ground-truth action chunk over the interval, in the benchmark's native action space ($7$-dim per-step end-effector control for CALVIN and VLABench, $12$-dim bimanual for RoboCOIN, and $16$-dim bimanual end-effector poses with grippers for AgiBot-Beta).

The three modalities are RGB $[\mathbf{I}_t, \mathbf{I}_{t+s}]$ (6 channels), RGB-D $[\mathbf{I}_t, \mathbf{D}_t, \mathbf{I}_{t+s}, \mathbf{D}_{t+s}]$ (8 channels), and RGB-DF, which further appends the dense optical flow $\mathbf{F}_{t\rightarrow t+s}$ estimated by DPFlow~\citep{morimitsu2025dpflow} (10 channels). Inputs are resized to $224\times224$ and fed into an ImageNet-pretrained ResNet-18 whose first convolution is widened to the corresponding channel count; the backbone is frozen and only the input convolution and the MLP action expert are trained with MSE loss on per-dimension z-normalized targets (statistics from the training split). We report the best validation MSE over 30 epochs; under this normalization a mean predictor scores $1.00$. All modalities are probed on identical frame pairs and splits, so differences reflect only the action-relevant information carried by the added channels.

\begin{table}[t]
  \caption{Results on the LARY benchmark~\citep{nie2026lary}. Representation probing results for low-level control regression (MSE, lower is better).}
 \label{tab:representation_probe}
  \centering
  \begin{tabular}{l ccccc}
    \toprule
    \textbf{Modality} & \textbf{Avg.}  & \textbf{CALVIN} & \textbf{VLABench} & \textbf{RoboCOIN} & \textbf{AgiBot-B} \\
    \midrule
    RGB    & 0.28 & 0.35 & 0.08 & 0.18 & 0.52 \\
    RGB-D  & 0.26 & 0.30 & 0.07 & 0.15 & 0.50 \\
    RGB-DF & \textbf{0.22} & \textbf{0.25} & \textbf{0.06} & \textbf{0.10} & \textbf{0.47} \\
    \bottomrule
  \end{tabular}
\end{table}

\subsection{Optical-flow Colorization and Decoding}
\label{app:flow-colorization}

Optical flow is a dense 2D vector field and does not have a natural RGB appearance.
In our pipeline, color-coded flow videos serve as the training representation for the flow branch, enabling it to reuse the pretrained RGB video VAE and standard video codecs.
This representation is also convenient for visualization and inspection.
However, color space is not a metric space for optical-flow accuracy.
Therefore, all quantitative flow metrics in this work are computed in flow-vector space after decoding the color-coded representations back to flow vectors.

For each pixel $p$, the flow vector is
\begin{equation}
\mathbf f_p = (u_p, v_p),
\end{equation}
where $u_p$ and $v_p$ denote horizontal and vertical displacements.
Since flow vectors do not have a natural RGB appearance, we use a Middlebury-style color-wheel visualization to convert flow fields into RGB videos.

\noindent
\textbf{Why color-coded flow videos?}
The color-coded representation is used for two practical reasons.
First, it provides an intuitive visualization: hue encodes flow direction, while brightness or saturation encodes flow magnitude.
Second, it allows flow sequences to be stored and processed using standard RGB video pipelines, which is convenient for video codecs, visualization, and pretrained video models.
However, this representation is not a metric space for optical-flow accuracy.
Therefore, quantitative flow evaluation should be performed in flow-vector space rather than in RGB color space.

\noindent
\textbf{Forward colorization.}
Given a flow vector $\mathbf f_p=(u_p,v_p)$, we compute its direction and magnitude:
\begin{equation}
\theta_p = \operatorname{atan2}(v_p, u_p),
\qquad
r_p = \sqrt{u_p^2 + v_p^2}.
\end{equation}
The direction $\theta_p$ is mapped to hue, and the normalized magnitude is mapped to value or saturation.
A common formulation is
\begin{equation}
H_p = \frac{\theta_p + \pi}{2\pi},
\qquad
S_p = 1,
\qquad
V_p = \min\left(\frac{r_p}{r_{\max}}, 1\right),
\end{equation}
where $r_{\max}$ is a normalization constant, such as the maximum flow magnitude in the clip or a robust percentile, e.g., the 95th percentile.
The RGB color is then obtained by HSV-to-RGB conversion:
\begin{equation}
I_p = \mathrm{HSV2RGB}(H_p, S_p, V_p).
\end{equation}
Invalid or occluded pixels are typically displayed as black.

\noindent
\textbf{Inverse decoding.}
In principle, if the colorization parameters are known, a color-coded flow image can be approximately decoded back to flow vectors.
Given an RGB pixel, we first convert it to HSV.
Assuming hue encodes direction and value encodes normalized magnitude, the flow direction and magnitude can be recovered as
\begin{equation}
\theta_p = 2\pi H_p - \pi,
\qquad
r_p = V_p \, r_{\max}.
\end{equation}
The flow vector is then
\begin{equation}
u_p = r_p \cos \theta_p,
\qquad
v_p = r_p \sin \theta_p.
\end{equation}
In practice, this decoding is approximate because video compression, color quantization, clipping, and unknown normalization can introduce errors. We therefore use the color-coded representation consistently throughout training and generation, and decode back to flow vectors only at evaluation time for metric computation.

\noindent
\textbf{Metric implication.}
Color-space distances between color-coded flow visualizations do not correspond to pixel displacement error and are therefore not used as evaluation metrics.
All endpoint-error metrics in this work are computed in flow-vector space using the definitions in Appendix~\ref{app:metrics-flow}.

%% file: paper.bib
@string{cvpr = "Proc. of IEEE Conf. on Computer Vision and Pattern Recognition"}

@string{iccv = "Proc. of IEEE Intl. Conf. on Computer Vision"}

@string{eccv = "Proc. of European Conf. on Computer Vision"}

@string{nips = "Proc. of Advances in Neural Information Processing Systems"}

@string{icml = "Proc. of Intl. Conf. on Machine Learning"}

@string{ijcai = "Proc. of Intl. Joint Conf. on Artificial Intelligence"}

@string{aaai = "Proc. of the AAAI Conf. on Artificial Intelligence"}

@article{ha2018recurrent,
  title={Recurrent world models facilitate policy evolution},
  author={Ha, David and Schmidhuber, J{\"u}rgen},
  journal=nips,
  volume={31},
  year={2018}
}

@article{xiang2024pandora,
  title={Pandora: Towards general world model with natural language actions and video states},
  author={Xiang, Jiannan and Liu, Guangyi and Gu, Yi and Gao, Qiyue and Ning, Yuting and Zha, Yuheng and Feng, Zeyu and Tao, Tianhua and Hao, Shibo and Shi, Yemin and others},
  journal={arXiv preprint arXiv:2406.09455},
  year={2024}
}

@article{zheng2024open,
  title={Open-sora: Democratizing efficient video production for all},
  author={Zheng, Zangwei and Peng, Xiangyu and Yang, Tianji and Shen, Chenhui and Li, Shenggui and Liu, Hongxin and Zhou, Yukun and Li, Tianyi and You, Yang},
  journal={arXiv preprint arXiv:2412.20404},
  year={2024}
}

@article{du2023learning,
  title={Learning universal policies via text-guided video generation},
  author={Du, Yilun and Yang, Sherry and Dai, Bo and Dai, Hanjun and Nachum, Ofir and Tenenbaum, Josh and Schuurmans, Dale and Abbeel, Pieter},
  journal=nips,
  volume={36},
  pages={9156--9172},
  year={2023}
}

@article{liang2024dreamitate,
  title={Dreamitate: Real-world visuomotor policy learning via video generation},
  author={Liang, Junbang and Liu, Ruoshi and Ozguroglu, Ege and Sudhakar, Sruthi and Dave, Achal and Tokmakov, Pavel and Song, Shuran and Vondrick, Carl},
  journal={arXiv preprint arXiv:2406.16862},
  year={2024}
}

@inproceedings{zhen2025learning,
  title={Learning 4D Embodied World Models},
  author={Zhen, Haoyu and Sun, Qiao and Zhang, Hongxin and Li, Junyan and Zhou, Siyuan and Du, Yilun and Gan, Chuang},
  booktitle=cvpr,
  pages={5337--5347},
  year={2025}
}

@article{zhu2024irasim,
  title={Irasim: Learning interactive real-robot action simulators},
  author={Zhu, Fangqi and Wu, Hongtao and Guo, Song and Liu, Yuxiao and Cheang, Chilam and Kong, Tao},
  journal={arXiv preprint arXiv:2406.14540},
  year={2024}
}

@article{du2023video,
  title={Video language planning},
  author={Du, Yilun and Yang, Mengjiao and Florence, Pete and Xia, Fei and Wahid, Ayzaan and Ichter, Brian and Sermanet, Pierre and Yu, Tianhe and Abbeel, Pieter and Tenenbaum, Joshua B and others},
  journal={arXiv preprint arXiv:2310.10625},
  year={2023}
}

@article{li2025novaflow,
  title={Novaflow: Zero-shot manipulation via actionable flow from generated videos},
  author={Li, Hongyu and Sun, Lingfeng and Hu, Yafei and Ta, Duy and Barry, Jennifer and Konidaris, George and Fu, Jiahui},
  journal={arXiv preprint arXiv:2510.08568},
  year={2025}
}

@article{ha2018world,
  title={World models},
  author={Ha, David and Schmidhuber, J{\"u}rgen},
  journal={arXiv preprint arXiv:1803.10122},
  year={2018}
}

@article{sutton1991dyna,
  title={Dyna, an integrated architecture for learning, planning, and reacting},
  author={Sutton, Richard S},
  journal={ACM Sigart Bulletin},
  volume={2},
  number={4},
  pages={160--163},
  year={1991},
  publisher={ACM New York, NY, USA}
}

@article{achille2018separation,
  title={A separation principle for control in the age of deep learning},
  author={Achille, Alessandro and Soatto, Stefano},
  journal={Annual Review of Control, Robotics, and Autonomous Systems},
  volume={1},
  number={1},
  pages={287--307},
  year={2018},
  publisher={Annual Reviews}
}

@article{lesort2018state,
  title={State representation learning for control: An overview},
  author={Lesort, Timoth{\'e}e and D{\'\i}az-Rodr{\'\i}guez, Natalia and Goudou, Jean-Franois and Filliat, David},
  journal={Neural Networks},
  volume={108},
  pages={379--392},
  year={2018},
  publisher={Elsevier}
}

@article{kong2024hunyuanvideo,
  title={Hunyuanvideo: A systematic framework for large video generative models},
  author={Kong, Weijie and Tian, Qi and Zhang, Zijian and Min, Rox and Dai, Zuozhuo and Zhou, Jin and Xiong, Jiangfeng and Li, Xin and Wu, Bo and Zhang, Jianwei and others},
  journal={arXiv preprint arXiv:2412.03603},
  year={2024}
}

@article{wang2025wan,
  title={Wan: Open and advanced large-scale video generative models},
  author={Wang, Ang and Ai, Baole and Wen, Bin and Mao, Chaojie and Xie, Chen-Wei and Chen, Di and Yu, Feiwu and Zhao, Haiming and Yang, Jianxiao and Zeng, Jianyuan and others},
  journal={arXiv preprint arXiv:2503.20314},
  year={2025}
}

@article{yang2024cogvideox,
  title={Cogvideox: Text-to-video diffusion models with an expert transformer},
  author={Yang, Zhuoyi and Teng, Jiayan and Zheng, Wendi and Ding, Ming and Huang, Shiyu and Xu, Jiazheng and Yang, Yuanming and Hong, Wenyi and Zhang, Xiaohan and Feng, Guanyu and others},
  journal={arXiv preprint arXiv:2408.06072},
  year={2024}
}

@article{chen20254dnex,
  title={4dnex: Feed-forward 4d generative modeling made easy},
  author={Chen, Zhaoxi and Liu, Tianqi and Zhuo, Long and Ren, Jiawei and Tao, Zeng and Zhu, He and Hong, Fangzhou and Pan, Liang and Liu, Ziwei},
  journal={arXiv preprint arXiv:2508.13154},
  year={2025}
}

@article{wang2021neus,
  title={Neus: Learning neural implicit surfaces by volume rendering for multi-view reconstruction},
  author={Wang, Peng and Liu, Lingjie and Liu, Yuan and Theobalt, Christian and Komura, Taku and Wang, Wenping},
  journal={arXiv preprint arXiv:2106.10689},
  year={2021}
}

@article{pfaff2020learning,
  title={Learning mesh-based simulation with graph networks},
  author={Pfaff, Tobias and Fortunato, Meire and Sanchez-Gonzalez, Alvaro and Battaglia, Peter W},
  journal={arXiv preprint arXiv:2010.03409},
  year={2020}
}

@inproceedings{jiang2025phystwin,
  title={Phystwin: Physics-informed reconstruction and simulation of deformable objects from videos},
  author={Jiang, Hanxiao and Hsu, Hao-Yu and Zhang, Kaifeng and Yu, Hsin-Ni and Wang, Shenlong and Li, Yunzhu},
  booktitle=cvpr,
  pages={7219--7230},
  year={2025}
}

@inproceedings{zhao2025physsplat,
  title={PhysSplat: Efficient Physics Simulation for 3D Scenes via MLLM-Guided Gaussian Splatting},
  author={Zhao, Haoyu and Wang, Hao and Zhao, Xingyue and Fei, Hao and Wang, Hongqiu and Long, Chengjiang and Zou, Hua},
  booktitle=iccv,
  pages={5242--5252},
  year={2025}
}

@inproceedings{xia2025drawer,
  title={Drawer: Digital reconstruction and articulation with environment realism},
  author={Xia, Hongchi and Su, Entong and Memmel, Marius and Jain, Arhan and Yu, Raymond and Mbiziwo-Tiapo, Numfor and Farhadi, Ali and Gupta, Abhishek and Wang, Shenlong and Ma, Wei-Chiu},
  booktitle=cvpr,
  pages={21771--21782},
  year={2025}
}

@inproceedings{xia2024video2game,
  title={Video2game: Real-time interactive realistic and browser-compatible environment from a single video},
  author={Xia, Hongchi and Lin, Zhi-Hao and Ma, Wei-Chiu and Wang, Shenlong},
  booktitle=cvpr,
  pages={4578--4588},
  year={2024}
}

@article{mildenhall2021nerf,
  title={Nerf: Representing scenes as neural radiance fields for view synthesis},
  author={Mildenhall, Ben and Srinivasan, Pratul P and Tancik, Matthew and Barron, Jonathan T and Ramamoorthi, Ravi and Ng, Ren},
  journal={Communications of the ACM},
  volume={65},
  number={1},
  pages={99--106},
  year={2021},
  publisher={ACM New York, NY, USA}
}

@article{kerbl20233d,
  title={3d gaussian splatting for real-time radiance field rendering.},
  author={Kerbl, Bernhard and Kopanas, Georgios and Leimk{\"u}hler, Thomas and Drettakis, George and others},
  journal={ACM Trans. Graph.},
  volume={42},
  number={4},
  pages={139--1},
  year={2023}
}

@inproceedings{driess2023learning,
  title={Learning multi-object dynamics with compositional neural radiance fields},
  author={Driess, Danny and Huang, Zhiao and Li, Yunzhu and Tedrake, Russ and Toussaint, Marc},
  booktitle={Conference on robot learning},
  pages={1755--1768},
  year={2023},
  organization={PMLR}
}

@inproceedings{xie2024physgaussian,
  title={Physgaussian: Physics-integrated 3d gaussians for generative dynamics},
  author={Xie, Tianyi and Zong, Zeshun and Qiu, Yuxing and Li, Xuan and Feng, Yutao and Yang, Yin and Jiang, Chenfanfu},
  booktitle=cvpr,
  pages={4389--4398},
  year={2024}
}

@inproceedings{sanchez2020learning,
  title={Learning to simulate complex physics with graph networks},
  author={Sanchez-Gonzalez, Alvaro and Godwin, Jonathan and Pfaff, Tobias and Ying, Rex and Leskovec, Jure and Battaglia, Peter},
  booktitle=icml,
  pages={8459--8468},
  year={2020},
  organization={PMLR}
}

@inproceedings{abou2024particlenerf,
  title={Particlenerf: A particle-based encoding for online neural radiance fields},
  author={Abou-Chakra, Jad and Dayoub, Feras and S{\"u}nderhauf, Niko},
  booktitle={Proceedings of the IEEE/CVF Winter Conference on Applications of Computer Vision},
  pages={5975--5984},
  year={2024}
}

@article{zhang2025particle,
  title={Particle-grid neural dynamics for learning deformable object models from rgb-d videos},
  author={Zhang, Kaifeng and Li, Baoyu and Hauser, Kris and Li, Yunzhu},
  journal={arXiv preprint arXiv:2506.15680},
  year={2025}
}

@inproceedings{kaelbling2011hierarchical,
  title={Hierarchical task and motion planning in the now},
  author={Kaelbling, Leslie Pack and Lozano-P{\'e}rez, Tom{\'a}s},
  booktitle={2011 IEEE international conference on robotics and automation},
  pages={1470--1477},
  year={2011},
  organization={IEEE}
}

@article{wang2025enact,
  title={Enact: Evaluating embodied cognition with world modeling of egocentric interaction},
  author={Wang, Qineng and Huang, Wenlong and Zhou, Yu and Yin, Hang and Bao, Tianwei and Lyu, Jianwen and Liu, Weiyu and Zhang, Ruohan and Wu, Jiajun and Fei-Fei, Li and others},
  journal={arXiv preprint arXiv:2511.20937},
  year={2025}
}

@article{zhao2026high,
  title={High-fidelity simulated data generation for real-world zero-shot robotic manipulation learning with gaussian splatting},
  author={Zhao, Haoyu and Zeng, Cheng and Zhuang, Linghao and Zhao, Yaxi and Xue, Shengke and Wang, Hao and Zhao, Xingyue and Li, Zhongyu and Li, Kehan and Huang, Siteng and others},
  journal={IEEE Robotics and Automation Letters},
  year={2026},
  publisher={IEEE}
}

@article{hu2024video,
  title={Video prediction policy: A generalist robot policy with predictive visual representations},
  author={Hu, Yucheng and Guo, Yanjiang and Wang, Pengchao and Chen, Xiaoyu and Wang, Yen-Jen and Zhang, Jianke and Sreenath, Koushil and Lu, Chaochao and Chen, Jianyu},
  journal={arXiv preprint arXiv:2412.14803},
  year={2024}
}

@article{williams2017model,
  title={Model predictive path integral control: From theory to parallel computation},
  author={Williams, Grady and Aldrich, Andrew and Theodorou, Evangelos A},
  journal={Journal of Guidance, Control, and Dynamics},
  volume={40},
  number={2},
  pages={344--357},
  year={2017},
  publisher={American Institute of Aeronautics and Astronautics}
}

@inproceedings{hafner2019learning,
  title={Learning latent dynamics for planning from pixels},
  author={Hafner, Danijar and Lillicrap, Timothy and Fischer, Ian and Villegas, Ruben and Ha, David and Lee, Honglak and Davidson, James},
  booktitle=icml,
  pages={2555--2565},
  year={2019},
  organization={PMLR}
}

@inproceedings{pineau2003point,
  title={Point-based value iteration: An anytime algorithm for POMDPs},
  author={Pineau, Joelle and Gordon, Geoff and Thrun, Sebastian and others},
  booktitle={Ijcai},
  volume={3},
  pages={1025--1032},
  year={2003}
}

@article{bi2025motus,
  title={Motus: A unified latent action world model},
  author={Bi, Hongzhe and Tan, Hengkai and Xie, Shenghao and Wang, Zeyuan and Huang, Shuhe and Liu, Haitian and Zhao, Ruowen and Feng, Yao and Xiang, Chendong and Rong, Yinze and others},
  journal={arXiv preprint arXiv:2512.13030},
  year={2025}
}

@article{hafner2019dream,
  title={Dream to control: Learning behaviors by latent imagination},
  author={Hafner, Danijar and Lillicrap, Timothy and Ba, Jimmy and Norouzi, Mohammad},
  journal={arXiv preprint arXiv:1912.01603},
  year={2019}
}

@article{hansen2023td,
  title={Td-mpc2: Scalable, robust world models for continuous control},
  author={Hansen, Nicklas and Su, Hao and Wang, Xiaolong},
  journal={arXiv preprint arXiv:2310.16828},
  year={2023}
}

@article{chua2018deep,
  title={Deep reinforcement learning in a handful of trials using probabilistic dynamics models},
  author={Chua, Kurtland and Calandra, Roberto and McAllister, Rowan and Levine, Sergey},
  journal=nips,
  volume={31},
  year={2018}
}

@article{hafner2025training,
  title={Training agents inside of scalable world models},
  author={Hafner, Danijar and Yan, Wilson and Lillicrap, Timothy},
  journal={arXiv preprint arXiv:2509.24527},
  year={2025}
}

@article{bharadhwaj2024gen2act,
  title={Gen2act: Human video generation in novel scenarios enables generalizable robot manipulation},
  author={Bharadhwaj, Homanga and Dwibedi, Debidatta and Gupta, Abhinav and Tulsiani, Shubham and Doersch, Carl and Xiao, Ted and Shah, Dhruv and Xia, Fei and Sadigh, Dorsa and Kirmani, Sean},
  journal={arXiv preprint arXiv:2409.16283},
  year={2024}
}

@article{ye2024latent,
  title={Latent action pretraining from videos},
  author={Ye, Seonghyeon and Jang, Joel and Jeon, Byeongguk and Joo, Sejune and Yang, Jianwei and Peng, Baolin and Mandlekar, Ajay and Tan, Reuben and Chao, Yu-Wei and Lin, Bill Yuchen and others},
  journal={arXiv preprint arXiv:2410.11758},
  year={2024}
}

@article{black2023zero,
  title={Zero-shot robotic manipulation with pretrained image-editing diffusion models},
  author={Black, Kevin and Nakamoto, Mitsuhiko and Atreya, Pranav and Walke, Homer and Finn, Chelsea and Kumar, Aviral and Levine, Sergey},
  journal={arXiv preprint arXiv:2310.10639},
  year={2023}
}

@inproceedings{brooks2023instructpix2pix,
  title={Instructpix2pix: Learning to follow image editing instructions},
  author={Brooks, Tim and Holynski, Aleksander and Efros, Alexei A},
  booktitle=cvpr,
  pages={18392--18402},
  year={2023}
}

@article{damen2020epic,
  title={The epic-kitchens dataset: Collection, challenges and baselines},
  author={Damen, Dima and Doughty, Hazel and Farinella, Giovanni Maria and Fidler, Sanja and Furnari, Antonino and Kazakos, Evangelos and Moltisanti, Davide and Munro, Jonathan and Perrett, Toby and Price, Will and others},
  journal={IEEE Transactions on Pattern Analysis and Machine Intelligence},
  volume={43},
  number={11},
  pages={4125--4141},
  year={2020},
  publisher={IEEE}
}

@article{wang2024egovid,
  title={Egovid-5m: A large-scale video-action dataset for egocentric video generation},
  author={Wang, Xiaofeng and Zhao, Kang and Liu, Feng and Wang, Jiayu and Zhao, Guosheng and Bao, Xiaoyi and Zhu, Zheng and Zhang, Yingya and Wang, Xingang},
  journal={arXiv preprint arXiv:2411.08380},
  year={2024}
}

@article{wu2024robomind,
  title={Robomind: Benchmark on multi-embodiment intelligence normative data for robot manipulation},
  author={Wu, Kun and Hou, Chengkai and Liu, Jiaming and Che, Zhengping and Ju, Xiaozhu and Yang, Zhuqin and Li, Meng and Zhao, Yinuo and Xu, Zhiyuan and Yang, Guang and others},
  journal={arXiv preprint arXiv:2412.13877},
  year={2024}
}

@article{wu2025robocoin,
  title={RoboCOIN: An Open-Sourced Bimanual Robotic Data COllection for INtegrated Manipulation},
  author={Wu, Shihan and Liu, Xuecheng and Xie, Shaoxuan and Wang, Pengwei and Li, Xinghang and Yang, Bowen and Li, Zhe and Zhu, Kai and Wu, Hongyu and Liu, Yiheng and others},
  journal={arXiv preprint arXiv:2511.17441},
  year={2025}
}

@article{liu2024rdt,
  title={Rdt-1b: a diffusion foundation model for bimanual manipulation},
  author={Liu, Songming and Wu, Lingxuan and Li, Bangguo and Tan, Hengkai and Chen, Huayu and Wang, Zhengyi and Xu, Ke and Su, Hang and Zhu, Jun},
  journal={arXiv preprint arXiv:2410.07864},
  year={2024}
}

@article{jiang2025galaxea,
  title={Galaxea open-world dataset and g0 dual-system vla model},
  author={Jiang, Tao and Yuan, Tianyuan and Liu, Yicheng and Lu, Chenhao and Cui, Jianning and Liu, Xiao and Cheng, Shuiqi and Gao, Jiyang and Xu, Huazhe and Zhao, Hang},
  journal={arXiv preprint arXiv:2509.00576},
  year={2025}
}

@article{bai2025qwen3,
  title={Qwen3-vl technical report},
  author={Bai, Shuai and Cai, Yuxuan and Chen, Ruizhe and Chen, Keqin and Chen, Xionghui and Cheng, Zesen and Deng, Lianghao and Ding, Wei and Gao, Chang and Ge, Chunjiang and others},
  journal={arXiv preprint arXiv:2511.21631},
  year={2025}
}

@article{lin2025depth,
  title={Depth anything 3: Recovering the visual space from any views},
  author={Lin, Haotong and Chen, Sili and Liew, Junhao and Chen, Donny Y and Li, Zhenyu and Shi, Guang and Feng, Jiashi and Kang, Bingyi},
  journal={arXiv preprint arXiv:2511.10647},
  year={2025}
}

@inproceedings{morimitsu2025dpflow,
  title={Dpflow: Adaptive optical flow estimation with a dual-pyramid framework},
  author={Morimitsu, Henrique and Zhu, Xiaobin and Cesar, Roberto M and Ji, Xiangyang and Yin, Xu-Cheng},
  booktitle=cvpr,
  pages={17810--17820},
  year={2025}
}

@article{lipman2022flow,
  title={Flow matching for generative modeling},
  author={Lipman, Yaron and Chen, Ricky TQ and Ben-Hamu, Heli and Nickel, Maximilian and Le, Matt},
  journal={arXiv preprint arXiv:2210.02747},
  year={2022}
}

@inproceedings{zhao2026towards,
  title={Towards affordance-aware robotic dexterous grasping with human-like priors},
  author={Zhao, Haoyu and Zhuang, Linghao and Zhao, Xingyue and Zeng, Cheng and Xu, Haoran and Jiang, Yuming and Cen, Jun and Wang, Kexiang and Guo, Jiayan and Huang, Siteng and others},
  booktitle=aaai,
  volume={40},
  number={15},
  pages={13126--13134},
  year={2026}
}

@article{zhao2025smap,
  title={Smap: Self-supervised motion adaptation for physically plausible humanoid whole-body control},
  author={Zhao, Haoyu and Lin, Sixu and Ben, Qingwei and Dai, Minyue and Fei, Hao and Wang, Jingbo and Zou, Hua and Dong, Junting},
  journal={arXiv preprint arXiv:2505.19463},
  year={2025}
}

@article{ren2024l4gm,
  title={L4gm: Large 4d gaussian reconstruction model},
  author={Ren, Jiawei and Xie, Kevin and Mirzaei, Ashkan and Liang, Hanxue and Zeng, Xiaohui and Kreis, Karsten and Liu, Ziwei and Torralba, Antonio and Fidler, Sanja and Kim, Seung W and others},
  journal=nips,
  volume={37},
  pages={56828--56858},
  year={2024}
}

@inproceedings{wu2025cat4d,
  title={Cat4d: Create anything in 4d with multi-view video diffusion models},
  author={Wu, Rundi and Gao, Ruiqi and Poole, Ben and Trevithick, Alex and Zheng, Changxi and Barron, Jonathan T and Holynski, Aleksander},
  booktitle=cvpr,
  pages={26057--26068},
  year={2025}
}

@article{zhao2024sg,
  title={Sg-gs: Photo-realistic animatable human avatars with semantically-guided gaussian splatting},
  author={Zhao, Haoyu and Yang, Chen and Wang, Hao and Zhao, Xingyue and Shen, Wei},
  journal={arXiv preprint arXiv:2408.09665},
  year={2024}
}

@article{zhao2024hfgs,
  title={Hfgs: 4d gaussian splatting with emphasis on spatial and temporal high-frequency components for endoscopic scene reconstruction},
  author={Zhao, Haoyu and Zhao, Xingyue and Zhu, Lingting and Zheng, Weixi and Xu, Yongchao},
  journal={arXiv preprint arXiv:2405.17872},
  year={2024}
}

@article{yu20244real,
  title={4real: Towards photorealistic 4d scene generation via video diffusion models},
  author={Yu, Heng and Wang, Chaoyang and Zhuang, Peiye and Menapace, Willi and Siarohin, Aliaksandr and Cao, Junli and Jeni, Laszlo A and Tulyakov, Sergey and Lee, Hsin-Ying},
  journal=nips,
  volume={37},
  pages={45256--45280},
  year={2024}
}

@inproceedings{bahmani20244d,
  title={4d-fy: Text-to-4d generation using hybrid score distillation sampling},
  author={Bahmani, Sherwin and Skorokhodov, Ivan and Rong, Victor and Wetzstein, Gordon and Guibas, Leonidas and Wonka, Peter and Tulyakov, Sergey and Park, Jeong Joon and Tagliasacchi, Andrea and Lindell, David B},
  booktitle=cvpr,
  pages={7996--8006},
  year={2024}
}

@inproceedings{wang2025continuous,
  title={Continuous 3d perception model with persistent state},
  author={Wang, Qianqian and Zhang, Yifei and Holynski, Aleksander and Efros, Alexei A and Kanazawa, Angjoo},
  booktitle=cvpr,
  pages={10510--10522},
  year={2025}
}

@inproceedings{li2025megasam,
  title={Megasam: Accurate, fast and robust structure and motion from casual dynamic videos},
  author={Li, Zhengqi and Tucker, Richard and Cole, Forrester and Wang, Qianqian and Jin, Linyi and Ye, Vickie and Kanazawa, Angjoo and Holynski, Aleksander and Snavely, Noah},
  booktitle=cvpr,
  pages={10486--10496},
  year={2025}
}

@article{black2024pi_0,
  title={{$\pi_0$}: A Vision-Language-Action Flow Model for General Robot Control},
  author={Black, Kevin and Brown, Noah and Driess, Danny and Esmail, Adnan and Equi, Michael and Finn, Chelsea and Fusai, Niccolo and Groom, Lachy and Hausman, Karol and Ichter, Brian and others},
  journal={arXiv preprint arXiv:2410.24164},
  year={2024}
}

@article{agarwal2025cosmos,
  title={Cosmos world foundation model platform for physical ai},
  author={Agarwal, Niket and Ali, Arslan and Bala, Maciej and Balaji, Yogesh and Barker, Erik and Cai, Tiffany and Chattopadhyay, Prithvijit and Chen, Yongxin and Cui, Yin and Ding, Yifan and others},
  journal={arXiv preprint arXiv:2501.03575},
  year={2025}
}

@article{li2026causal,
  title={Causal World Modeling for Robot Control},
  author={Li, Lin and Zhang, Qihang and Luo, Yiming and Yang, Shuai and Wang, Ruilin and Han, Fei and Yu, Mingrui and Gao, Zelin and Xue, Nan and Zhu, Xing and others},
  journal={arXiv preprint arXiv:2601.21998},
  year={2026}
}

@article{ali2025world,
  title={World simulation with video foundation models for physical ai},
  author={Ali, Arslan and Bai, Junjie and Bala, Maciej and Balaji, Yogesh and Blakeman, Aaron and Cai, Tiffany and Cao, Jiaxin and Cao, Tianshi and Cha, Elizabeth and Chao, Yu-Wei and others},
  journal={arXiv preprint arXiv:2511.00062},
  year={2025}
}

@article{ye2026world,
  title={World action models are zero-shot policies},
  author={Ye, Seonghyeon and Ge, Yunhao and Zheng, Kaiyuan and Gao, Shenyuan and Yu, Sihyun and Kurian, George and Indupuru, Suneel and Tan, You Liang and Zhu, Chuning and Xiang, Jiannan and others},
  journal={arXiv preprint arXiv:2602.15922},
  year={2026}
}

@inproceedings{liu2025free4d,
  title={Free4d: Tuning-free 4d scene generation with spatial-temporal consistency},
  author={Liu, Tianqi and Huang, Zihao and Chen, Zhaoxi and Wang, Guangcong and Hu, Shoukang and Shen, Liao and Sun, Huiqiang and Cao, Zhiguo and Li, Wei and Liu, Ziwei},
  booktitle=cvpr,
  pages={25571--25582},
  year={2025}
}

@article{chi2025diffusion,
  title={Diffusion policy: Visuomotor policy learning via action diffusion},
  author={Chi, Cheng and Xu, Zhenjia and Feng, Siyuan and Cousineau, Eric and Du, Yilun and Burchfiel, Benjamin and Tedrake, Russ and Song, Shuran},
  journal={The International Journal of Robotics Research},
  volume={44},
  number={10-11},
  pages={1684--1704},
  year={2025},
  publisher={Sage Publications Sage UK: London, England}
}

@article{zhou2025pai,
  title={PAI-Bench: A Comprehensive Benchmark For Physical AI},
  author={Zhou, Fengzhe and Huang, Jiannan and Li, Jialuo and Ramanan, Deva and Shi, Humphrey},
  journal={arXiv preprint arXiv:2512.01989},
  year={2025}
}

@article{zhang2022dino,
  title={Dino: Detr with improved denoising anchor boxes for end-to-end object detection},
  author={Zhang, Hao and Li, Feng and Liu, Shilong and Zhang, Lei and Su, Hang and Zhu, Jun and Ni, Lionel M and Shum, Heung-Yeung},
  journal={arXiv preprint arXiv:2203.03605},
  year={2022}
}

@article{intelligence2025pi_,
  title={{{$\pi_{0.5}$}}: A Vision-Language-Action Model with Open-World Generalization},
  author={Intelligence, Physical and Black, Kevin and Brown, Noah and Darpinian, James and Dhabalia, Karan and Driess, Danny and Esmail, Adnan and Equi, Michael and Finn, Chelsea and Fusai, Niccolo and others},
  journal={arXiv preprint arXiv:2504.16054},
  year={2025}
}

@inproceedings{ke2021musiq,
  title={Musiq: Multi-scale image quality transformer},
  author={Ke, Junjie and Wang, Qifei and Wang, Yilin and Milanfar, Peyman and Yang, Feng},
  booktitle=iccv,
  pages={5148--5157},
  year={2021}
}

@inproceedings{li2023amt,
  title={Amt: All-pairs multi-field transforms for efficient frame interpolation},
  author={Li, Zhen and Zhu, Zuo-Liang and Han, Ling-Hao and Hou, Qibin and Guo, Chun-Le and Cheng, Ming-Ming},
  booktitle=cvpr,
  pages={9801--9810},
  year={2023}
}

@inproceedings{esser2024scaling,
  title={Scaling rectified flow transformers for high-resolution image synthesis},
  author={Esser, Patrick and Kulal, Sumith and Blattmann, Andreas and Entezari, Rahim and M{\"u}ller, Jonas and Saini, Harry and Levi, Yam and Lorenz, Dominik and Sauer, Axel and Boesel, Frederic and others},
  booktitle=icml,
  year={2024}
}

@inproceedings{he2016deep,
  title={Deep residual learning for image recognition},
  author={He, Kaiming and Zhang, Xiangyu and Ren, Shaoqing and Sun, Jian},
  booktitle=cvpr,
  pages={770--778},
  year={2016}
}

@article{bu2025agibot,
  title={Agibot world colosseo: A large-scale manipulation platform for scalable and intelligent embodied systems},
  author={Bu, Qingwen and Cai, Jisong and Chen, Li and Cui, Xiuqi and Ding, Yan and Feng, Siyuan and Gao, Shenyuan and He, Xindong and Hu, Xuan and Huang, Xu and others},
  journal={arXiv preprint arXiv:2503.06669},
  year={2025}
}

@article{guo2026articulat3d,
  title={Articulat3D: Reconstructing Articulated Digital Twins From Monocular Videos with Geometric and Motion Constraints},
  author={Guo, Lijun and Zhao, Haoyu and Zhao, Xingyue and Fu, Rong and Zhuang, Linghao and Huang, Siteng and Li, Zhongyu and Zou, Hua},
  journal={arXiv preprint arXiv:2603.11606},
  year={2026}
}

@article{pai2025mimic,
  title={mimic-video: Video-action models for generalizable robot control beyond vlas},
  author={Pai, Jonas and Achenbach, Liam and Montesinos, Victoriano and Forrai, Benedek and Mees, Oier and Nava, Elvis},
  journal={arXiv preprint arXiv:2512.15692},
  year={2025}
}

@inproceedings{nie2026lary,
      title={LARY: A Latent Action Representation Yielding Benchmark for Generalizable Vision-to-Action Alignment},
      author={Dujun Nie and Fengjiao Chen and Qi Lv and Jun Kuang and Xiaoyu Li and Xuezhi Cao and Xunliang Cai},
      booktitle={European Conference on Computer Vision (ECCV)},
      year={2026}
}

@article{zhao2026rynnworld,
  title={RynnWorld-Teleop: An Action-Conditioned World Model for Digital Teleoperation},
  author={Zhao, Haoyu and Zhao, Xingyue and Li, Hangyu and Gong, Biao and Li, Kehan and Huang, Siteng and Li, Xin and Zhao, Deli and Li, Zhongyu},
  journal={arXiv preprint arXiv:2607.06558},
  year={2026}
}
